\documentclass[10pt]{article}
\usepackage[letterpaper,margin=1in]{geometry}
\usepackage{times}
\usepackage[round,authoryear]{natbib}
\usepackage[T1]{fontenc}
\usepackage{amsmath,amssymb,booktabs,graphicx}
\usepackage{microtype}
\usepackage[hidelinks]{hyperref}
\usepackage{url}
\newtheorem{proposition}{Proposition}

\newcommand{\PBB}{\textsc{PBB}}
\newcommand{\Full}{\mathrm{Full}}
\newcommand{\SWA}{\mathrm{SWA}}
\newcommand{\A}{\mathcal{A}}
\newcommand{\R}{\mathcal{R}}
\newcommand{\ind}{\mathbf{1}}
\title{\bfseries Post-Boundary Bridge:\\Must Local Attention Go Global Between Global Layers?}
\author{Zhibo Yang\\
Ocean University of China\\
\texttt{yzbdeeplearning@163.com}}

\date{September 2026}
\begin{document}
\raggedbottom
\maketitle
\begin{abstract}
Hybrid Transformers reduce the cost of long-context modeling by combining local attention with periodic full-attention layers. When global communication is already available, however, the best use of local computation remains unclear. We introduce Post-Boundary Bridge (PBB), which preserves causal attention within blocks and adds direct connections across their boundaries. Rather than extending the range of information relayed through successive local layers, PBB prioritizes within-block modeling and nearby exchange, leaving long-range communication to full-attention layers. Across dense and mixture-of-experts models with 205 million to 2.07 billion stored parameters, PBB hybrids retain near-Full perplexity and competitive performance on standard downstream benchmarks while improving controlled source retrieval. At 205 million parameters, PBB also matches a hybrid using sliding-window attention (SWA) in perplexity and achieves higher source-retrieval accuracy. The same boundary-aligned structure enables Flash-PBB, an exact decoding implementation that updates its key--value cache without relocating retained entries. Compared with Flash-SWA, Flash-PBB delivers $1.82\times$ decode attention-core throughput with half the allocated local key--value cache. These results show that targeted boundary exchange can preserve model quality while enabling faster, more memory-efficient local attention between global layers.
\end{abstract}

\begin{figure}[ht]
 \centering\includegraphics[width=\linewidth]{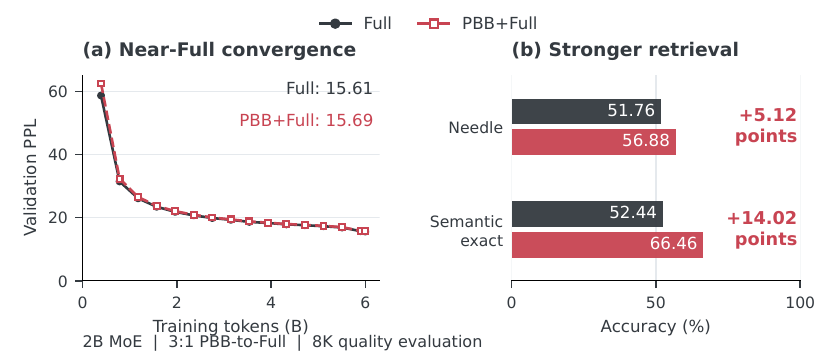}
 \caption{\textbf{Near-Full convergence, stronger retrieval.} 2B MoE. (a) All 16 logged validation points, unsmoothed. (b) Literal needle and Semantic exact retrieval: 6,720 balanced examples per task, 25\% chance accuracy. Both quality evaluations use 8K context.}
 \label{fig:overview}
\end{figure}

\section{Introduction}
Full attention gives every token direct access to its causal prefix, but prefill computes a quadratic number of scores and decode repeatedly reads an expanding KV history. Hybrid architectures reduce these costs by interleaving local attention, commonly sliding-window attention (SWA), with global layers \citep{gemmateam2024gemma2}. This changes the local layer's role: it no longer has to provide all communication by itself. Yet a local pattern that works well alone is not necessarily the best use of computation between global layers.

SWA preserves recent history while extending theoretical reach through cross-layer relay. Production-scale evidence motivates distinguishing these functions: DeepSeek-V4.1-Flash reports little quality loss when bounded replay reconstructs recent SWA states while approximating earlier local propagation and retaining cached global information \citep{deepseekai2026v41flash}. We ask an architectural question suggested by this observation: \emph{when periodic Full layers already provide global communication, can a smaller local graph preserve the nearby exchanges that matter?}

Block attention is a simple starting point, but its boundaries sever even adjacent-token dependencies. Stacking fixed blocks cannot repair this barrier; in a hybrid, newly computed cross-block features must wait for the next Full layer. We introduce \emph{Post-Boundary Bridge} (\PBB) to restore immediate exchange: each target keeps its causal block prefix, and the first half-block also reads the preceding half-block. One softmax combines these sources, without extra parameters or routing. PBB retains SWA's one-layer exchange for these boundary pairs using 24.4\% fewer local scores. Our hybrid repeats three PBB layers and one Full layer, with the same grouped-query attention (GQA) projections as Full \citep{ainslie2023gqa}.

\textbf{Stronger retrieval with near-Full modeling quality.} We evaluate 205M/352M-parameter dense models (\emph{200M}/\emph{350M}) and a 2.07B-stored/300M-active mixture-of-experts model (\emph{2B MoE}). The MoE hybrid improves controlled needle and Semantic exact retrieval by 5.12 and 14.02 points while closely tracking Full's convergence (Figure~\ref{fig:overview}). Both tasks improve in accuracy and source sensitivity across the evaluated Full/PBB comparisons. At 200M, PBB+Full lowers perplexity (PPL) and improves balanced source retrieval by 11.5--11.8 points. The SWA comparison reveals the architectural distinction: pure SWA outperforms pure PBB, but PBB+Full matches SWA+Full's PPL and improves balanced source retrieval by 4.81 points. The best standalone local pattern need not be the best complement to global attention.

\textbf{The same support also enables a faster, smaller local operator.} Flash-PBB maps the boundary-aligned sources to a fixed ring, fusing KV append and attention without relocating retained entries. It delivers $1.82\times$ Flash-SWA's decode-core throughput with half the allocated local KV. Profiling shows fewer registers, less shared memory, and sharply fewer memory requests (Figure~\ref{fig:flash_decode}). At whole-model scope, the MoE hybrid reaches $3.32\times$ Full's prefill and $2.11\times$ decode throughput at 512K, and $1.72\times$ training-update throughput at 64K (Figure~\ref{fig:efficiency}).

Our contributions are: (i) a boundary-repair design grounded in the distinction between global input coverage and timely local exchange; (ii) controlled evidence that the preferred local pattern changes with the layer schedule, together with retrieval gains across dense and MoE families; and (iii) Flash-PBB, an exact decode implementation that converts this support into relocation-free caching and measured core and whole-model gains.

\section{Local Exchange Between Global Layers}
\label{sec:theory}
\subsection{Static topology as a design choice}
Static attention offers deterministic access and predictable cache requirements without query-dependent routing. Sliding windows serve Mistral and Gemma 2 \citep{jiang2023mistral,gemmateam2024gemma2} with mature exact kernels \citep{dao2022flashattention,dao2024flashattention2}, making static topology a practical setting for studying edge allocation.

We distinguish \emph{which inputs are reachable} from \emph{how soon a fresh feature can move}. Full layers supply global input coverage; the intervening local graph determines next-layer boundary exchange.

\subsection{Structural dependency closure}
For a length-$n$ sequence, positions are integers in $\{0,\ldots,n-1\}$. Define the causal prefix $\mathcal C_t=\{s:0\leq s\leq t\}$ and
\begin{equation}
 p(t)=b\lfloor t/b\rfloor,\qquad r(t)=t-p(t),\qquad
 B_j=\{jb,\ldots,\min((j+1)b,n)-1\}.
\end{equation}
Here $b$ is block size, $p(t)$ is the block start, and $r(t)$ is the target phase. Let $\A_\ell(t)\subseteq\mathcal C_t$ be the sources read at $t$ in layer $\ell$. With positionwise non-attention operations, structural dependency closure is
\begin{equation}
 \R_0(t)=\{t\},\qquad
 \R_{\ell+1}(t)=\R_\ell(t)\cup\bigcup_{s\in\A_{\ell+1}(t)}\R_\ell(s).
 \label{eq:closure}
\end{equation}
The first term includes the residual path. Agreement on $\R_\ell(t)$ suffices for agreement of the layer-$\ell$ representation at $t$. Learned parameters need not use every allowed path, but cannot create a path outside this closure.

\begin{proposition}[Fixed-block barrier]
If every layer uses the same partition and $\A_{\rm Block}(t)=B_{\lfloor t/b\rfloor}\cap\mathcal C_t$, then $\R_L(t)\subseteq\{p(t),\ldots,t\}$ for every depth $L$. At a block boundary $p$, position $p-1$ cannot influence position $p$.
\label{prop:barrier}
\end{proposition}
For a constructed $K$-way boundary-copy distribution, let a uniform label be revealed only at $p-1$, independently of the accessible target prefix. A fixed-block readout at $p$ then has top-1 accuracy at most $1/K$ and expected cross-entropy at least $\log K$. Adding the direct edge removes the structural obstruction. Appendix~\ref{app:proofs} gives the proofs and independence conditions.

\begin{proposition}[Sliding-window relay]
For $\A_{\SWA}(t)=\{\max(0,t-w+1),\ldots,t\}$, where window size $w$ includes the current position, $L$ identical local layers give
\begin{equation}
 \R_L(t)=\{\max(0,t-L(w-1)),\ldots,t\}.
 \label{eq:swa_relay}
\end{equation}
After a Full layer, the closure equals $\mathcal C_t$ and remains so under subsequent residual causal layers.
\label{prop:relay}
\end{proposition}
\paragraph{Why a boundary bridge still matters after Full.}
Let $\tau_\ell(s,t)$ count the additional layers needed for a perturbation introduced at position $s$ after layer $\ell$ to reach $t$, with other layer-$\ell$ states held fixed. If $f$ is the next Full layer, then for the tail-to-head pairs repaired by PBB,
\begin{equation}
 \tau_\ell^{\mathrm{Block+Full}}(s,t)=f-\ell,\qquad
 \tau_\ell^{\mathrm{PBB+Full}}(s,t)=\tau_\ell^{\mathrm{SWA+Full}}(s,t)=1.
 \label{eq:latency_main}
\end{equation}
For the 3:1 schedule, this replaces a wait of up to four layers with next-layer exchange. An earlier Full layer provides input coverage, but has not broadcast this fresh feature. Appendix~\ref{app:local_value} proves the statement. The design objective is therefore to preserve short boundary paths, not to maximize every local relay path. Learned modeling quality tests this allocation empirically.

\section{Post-Boundary Bridge Attention}
\label{sec:method}
\begin{figure}[t]
 \centering\includegraphics[width=\linewidth]{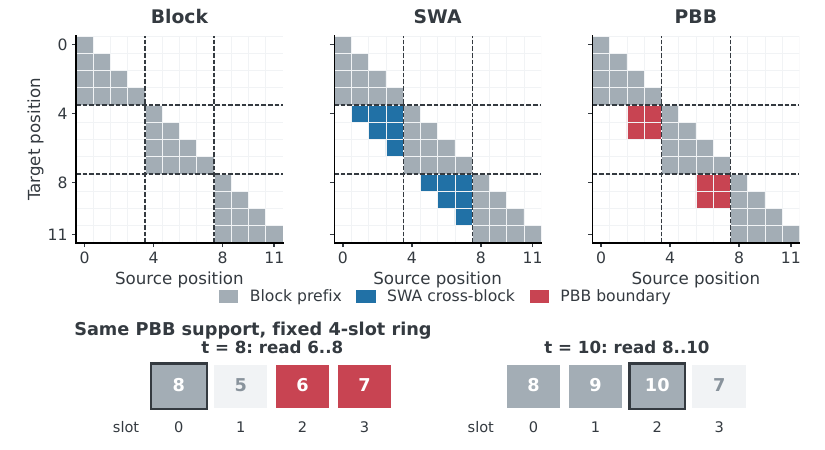}
 \caption{\textbf{One support, two benefits: direct exchange and relocation-free storage.} Top: block-prefix edges (gray) and cross-block edges (blue/red). Bottom: the same PBB sources in a fixed ring; numbers are absolute token positions, pale entries are expired. Only the newly appended slot is written. Illustration: $b=w=4$, $h=2$; models: $b=w=128$, $h=64$, every fourth layer Full.}
 \label{fig:method}
\end{figure}

Write a target as $t=p+r$. PBB keeps the causal block prefix and repairs the boundary for the first $h$ target phases:
\begin{equation}
 \A_{\PBB}(t)=\{p,\ldots,t\}\ \cup\
 \begin{cases}
  \{p-h,\ldots,p-1\},&p>0,\ 0\leq r<h,\\
  \varnothing,&\text{otherwise}.
 \end{cases}
 \label{eq:pbb}
\end{equation}
We use $b=128$, $h=64$, and SWA window $w=128$. In a centered auxiliary window, only post-boundary targets acquire new causal sources; pre-boundary targets already read those sources through their block prefix. PBB keeps the nonredundant edges with no additional projections, gates, or branch scales. Every fourth layer is Full; Figure~\ref{fig:method} connects the local geometry to its cache layout.

For query $q_t$, key $k_s$, value $v_s$, and head dimension $d_h$, define $z_{ts}=q_t^\top k_s/\sqrt{d_h}$. The output before the shared output projection is
\begin{equation}
 o_t=\frac{\sum_{s\in\A_{\PBB}(t)}e^{z_{ts}}v_s}{\sum_{s\in\A_{\PBB}(t)}e^{z_{ts}}}.
 \label{eq:attention}
\end{equation}
All main PBB models use this single softmax. A fused Triton kernel processes disjoint block and bridge tiles with log-sum-exp rescaling (Appendix~\ref{app:proofs}), native GQA, and backward propagation. Flash-PBB specializes cached decode to the same support. Neither repeats KV heads or materializes a dense mask. Historical separately normalized branches are distinguished in Appendix~\ref{app:legacy}.

\paragraph{Flash-PBB: exact attention in a fixed ring.}
With $b=2h$, the same support is the interval $[a(t),t]$, where $a(t)=\max(0,h(\lfloor t/h\rfloor-1))$. It spans at most $b$ positions, and its left endpoint advances only at half-block boundaries. Store token $s$ at slot $s\bmod b$: the overwritten token is already outside the support and can never be needed again. This gives an exact $b$-slot cache without relocating retained entries (proof in Appendix~\ref{app:ring_cache}). Flash-PBB fuses KV append and attention over this 128-slot ring, using a 128-column tile and native GQA. The kernel reads wrapped sources directly, eliminating staging gathers and rollover copies.

\paragraph{Coverage and arithmetic.}
For noninitial blocks, direct coverage at source distance $d=t-s$ and target phase $r$ is
\begin{equation}
 C_{\PBB}(d,r)=\ind\{0\leq d\leq r\}+\ind\{r<h,\ r<d\leq r+h\},
\end{equation}
where $\ind$ denotes an indicator. With $w=b=2h$, PBB's direct support is a subset of SWA's. PBB also relays information across layers: each additional local layer admits one more preceding half-block (Appendix~\ref{app:proofs}). Its design prioritizes boundary exchange within a smaller local edge budget.

At 8K, PBB uses 786,432 scores per head and local layer versus SWA's 1,040,448. It retains every within-block score and approximately half the cross-block scores: 258,048 versus 512,064. The 20-layer hybrid uses 26.76\% of all-Full's attention scores; with fixed local widths this fraction approaches one quarter as context grows. Only five Full layers retain the unbounded KV history. The next sections test both consequences: cheaper execution and useful source exchange.

\section{Experimental Setup}
\label{sec:setup}
We train Qwen2.5-like decoders \citep{qwen2025qwen25} on filtered FineWeb-Edu \citep{penedo2024fineweb}, using the Qwen2.5 tokenizer, rotary positions, root-mean-square normalization (RMSNorm), tied embeddings, 20 layers, two KV heads, and 8K sequences. The MoE uses 64 routed experts, top-4 routing, and a shared expert. Names ending in ``+Full'' denote the 3:1 hybrid; other local names denote 20 local layers.

Pairs match parameterization, data order, exposure, optimizer, and schedule. Updates contain 393,216 tokens; budgets are 4.109B/7.078B/6.000B for 200M/350M/MoE, approximately 20 per dense or active parameter. AdamW uses peak/minimum rates $3\times10^{-4}/3\times10^{-5}$, 5.56\% warmup, a plateau to 88.89\%, then cosine decay. We compare final checkpoints and retain 5,000-step and best-validation checkpoints. Run configurations and result provenance are specified in Appendix~\ref{app:recipe}.

\paragraph{Quality metrics.}
Quality evaluation uses models trained at 8K context; longer-context benchmarks measure computational efficiency. We measure held-out PPL, negative log-likelihood (NLL) by block phase, and zero-shot likelihood tasks using the LM Evaluation Harness \citep{gao2023harness}. \emph{LM-6} is the unweighted mean accuracy of LAMBADA, HellaSwag, PIQA, ARC-Easy, SciQ, and Winogrande. BLiMP \citep{warstadt2020blimp} and ARC-Challenge are reported separately.

The structural probe retrieves a supplied source token from four one-token candidates. Matched groups rotate all four labels to separate source use from candidate priors. Fifteen distances, seven readout phases, and 16 groups per cell yield 1,680 groups/6,720 examples, with 25\% chance accuracy. \emph{Probe} is candidate accuracy; \emph{source gain} is the correct label's mean log-probability increase under its supporting source versus counterfactual sources, in nats. These behavioral metrics differ from structural reachability. Prompts are shared and frozen after Full-only calibration (Appendix~\ref{app:probes}). Literal needle and Semantic exact retrieval each add 6,720 balanced examples on a separate distance grid (Appendix~\ref{app:retrieval_families}).

\section{Efficiency: Flash-PBB versus Flash-SWA}
\label{sec:efficiency}
The smaller graph must save more than attention scores: its execution must also avoid cache movement and oversized tiles. We test Flash-PBB against Flash-SWA at the local core, then measure the complete hybrid against Full. Timings use RTX 4090 (48\,GB) GPUs, FlashAttention 2.8.3.post1, synthetic inputs, and matched GQA, precision, expert backends, and phase-specific settings (Appendix~\ref{app:efficiency}).

\paragraph{$1.82\times$ core throughput with half the local KV.}
Across the 200M, 350M, and 2B attention shapes and eight position origins, Flash-PBB delivers $1.82\times$ Flash-SWA's decode-core throughput, including KV maintenance. At the 2B shape, time falls from 8.512 to 4.672\,$\mu$s/token, a 45.1\% reduction. Its 128-slot ring halves allocated local KV from 128 to 64\,KiB per layer and eliminates rollover copies. All 24 shape/position comparisons use the same device and three independent runs (Appendix~\ref{app:flash_decode}).

\begin{figure}[t]
 \centering\includegraphics[width=\linewidth]{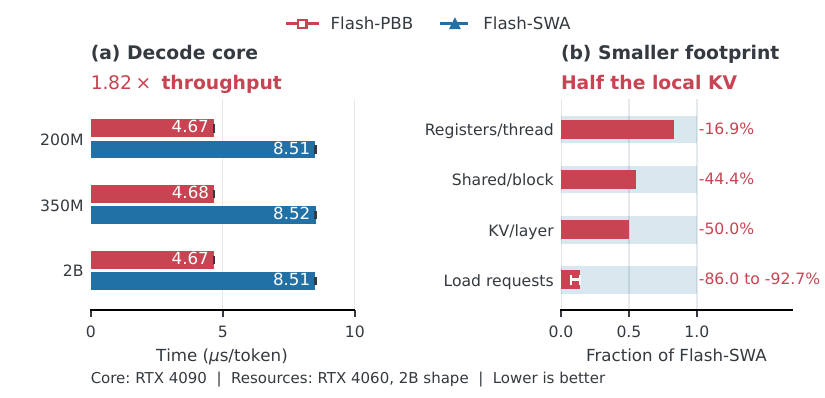}
 \caption{\textbf{Flash-PBB: faster local decoding with a smaller footprint.} (a) Three attention shapes, 8K position origin, RTX 4090 (48\,GB): median and range over three runs. (b) RTX 4060, 2B shape, normalized to Flash-SWA. White marks span six phases of L1/TEX global-load sector requests; other resource counts are phase-invariant.}
 \label{fig:flash_decode}
\end{figure}

\paragraph{Fewer execution resources and memory requests.}
The implementation reduces the work surrounding the dot products as well. Profiling the 2B attention shape shows 212 versus 255 registers/thread (16.9\% fewer) and 40 versus 72\,KiB dynamic shared memory/block (44.4\% less). Global-load sector requests fall from 15,168 to 1,112--2,120 per invocation, 86.0--92.7\% fewer across six phases; thread-local load/store requests fall from 2,880/5,952 to zero. These RTX 4060 counts repeat across three runs and two cache policies. Figure~\ref{fig:flash_decode} pairs the core speedup with these resource measurements: the regular support is useful both as a sparse graph and as a compact GPU working set (Appendix~\ref{app:decode_resources}).

\begin{figure}[t]
 \centering\includegraphics[width=\linewidth]{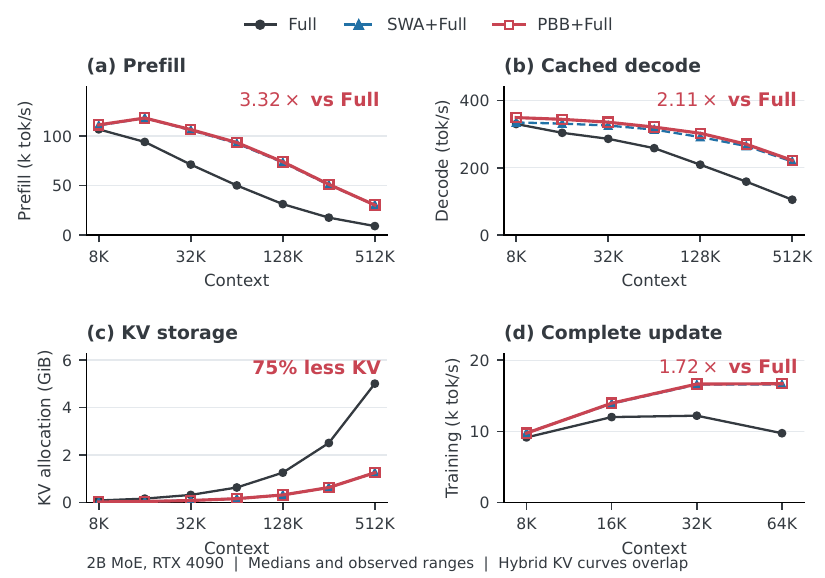}
 \caption{\textbf{Whole-model efficiency with matched optimized backends.} (a) Cache-populating prefill, three warmed calls. (b) Sequential cached decode, three 256-token CUDA-Graph continuations. (c) Physical KV allocation; hybrid curves nearly overlap. (d) Complete training updates, five after two warmups at 131,072 tokens/update. Curves show medians and ranges under phase-specific protocols. All successful lengths are shown; 1M inference and training at 128K or above exhaust memory. $K=1024$; k tok/s uses $k=1000$.}
 \label{fig:efficiency}
\end{figure}

\paragraph{Whole-model gains against Full.}
At 512K, the 2B MoE hybrid reaches 30.31k versus Full's 9.13k prefill tokens/s ($3.32\times$) and 221.87 versus 105.11 sequential decode tokens/s ($2.11\times$). Timing includes projections, experts, KV updates, and token selection; CUDA Graphs pass 384-step eager logit checks. Persistent KV falls from 5.004 to 1.252\,GiB (75.0\% less), as only five Full layers retain the long history.

Complete training-update throughput reaches 16.70k versus 9.74k tokens/s at 64K ($1.72\times$). The sweep fixes micro-batch 1 and 131,072 tokens/update, with BF16 computation, FP32 master states, and clearing, forward, backward, clipping, and AdamW included. Appendix~\ref{app:efficiency} reports all timings and capacity attempts.

SWA+Full reaches 30.39k prefill and 218.34 decode tokens/s at 512K. These end-to-end rates include Full layers, projections, and experts; the $1.82\times$ gain concerns the local core.

The remaining question is whether these savings preserve useful information exchange. The following comparisons isolate the effects of global layers, edge placement, and model scale.

\section{Stronger Retrieval with Targeted Boundary Exchange}
\label{sec:200m}
\subsection{Stronger source use with less local computation}
\begin{table}[t]
 \centering\small\begin{tabular}{llrrrr}
\toprule
Run & Architecture & PPL & Probe (\%) & Source gain & LM-6 (\%) \\
\midrule
A & Full & 21.52 & 69.84 & 1.645 & 45.78 \\
A & PBB+Full & 21.27 & 81.31 & 2.543 & 46.13 \\
B & Full & 21.38 & 68.26 & 1.811 & 46.55 \\
B & PBB+Full & 21.22 & 80.07 & 2.405 & 45.98 \\
\bottomrule
\end{tabular}

 \caption{\textbf{Full versus PBB+Full at matched training exposure.} Both architectures use GQA and 4.109B training tokens. PBB+Full has 15 local and five Full layers. Probe and LM-6 are accuracies in percent; source gain is in nats. Run identifiers are defined in Appendix~\ref{app:recipe}.}
 \label{tab:seeds}
\end{table}
\textbf{Does replacing Full sacrifice source use?} Replacing 15 of 20 Full layers instead improves retrieval and lowers PPL in the matched comparisons, at 26.76\% of Full's attention scores (Table~\ref{tab:seeds}). Probe gains of 11.47/11.82 points have paired group-bootstrap 95\% intervals $[10.46,12.50]$/$[10.74,12.95]$. Source gain rises by 0.898/0.594 nats, linking the accuracy increase to stronger sensitivity to the supplied source. The advantage persists at 5k and 10k updates (Appendix~\ref{app:probe_checkpoints}) and on separate needle and Semantic exact tasks, which gain 3.42/6.21 and 4.75/3.71 points (Appendix~\ref{app:retrieval_families}). BLiMP improves by 1.77/1.46 points with paired intervals excluding zero; ARC-Challenge also improves. LM-6 has mixed changes across runs (Appendix~\ref{app:downstream_paired}).

\subsection{The best local pattern depends on the schedule}
\begin{table}[t]
 \centering\small\begin{tabular}{lrrrr}
\toprule
Architecture & PPL $\downarrow$ & Probe $\uparrow$ & Source gain $\uparrow$ & LM-6 $\uparrow$ \\
\midrule
Full & 21.52 & 69.84 & 1.645 & 45.78 \\
Block & 30.42 & 35.83 & 0.315 & 44.85 \\
SWA & 23.31 & 54.27 & 0.911 & \textbf{46.13} \\
PBB & 24.05 & 46.80 & 0.647 & 44.95 \\
Random-Edge & 25.63 & 42.02 & 0.576 & 45.27 \\
SWA+Full & 21.29 & 76.50 & 2.305 & 45.82 \\
WideBlock-192+Full & 21.67 & 80.57 & 2.336 & 45.92 \\
PBB+Full & \textbf{21.27} & \textbf{81.31} & \textbf{2.543} & \textbf{46.13} \\
\bottomrule
\end{tabular}

 \caption{\textbf{Local-family and equal-arithmetic controls at 200M.} Pure PBB/Random-Edge and hybrid PBB/WideBlock-192 match their respective attention arithmetic and non-attention modules. Block's aggregate includes reachable and unreachable groups. Metric definitions match Table~\ref{tab:seeds}.}
 \label{tab:200m}
\end{table}
\textbf{Does the preferred local graph change when Full layers are added?} With all 20 layers local, SWA has lower PPL (23.31 versus 24.05) and higher probe accuracy (54.27\% versus 46.80\%) than PBB. With five Full layers, the hybrids reach comparable PPL (21.291 for SWA+Full; 21.266 for PBB+Full), while the probe ordering reverses: PBB+Full reaches 81.31\% versus 76.50\%. Its paired advantage is 4.81 points, with interval $[3.96,5.62]$; source gain increases by 0.238 nats, with interval $[0.203,0.273]$.

PBB's support is a subset of SWA's, yet the broader graph is not the stronger hybrid on this probe (Figure~\ref{fig:ordering}). This contrast answers our design question: choosing a local component in isolation can miss its value alongside global attention. SWA is stronger when local layers carry all communication; targeted boundary exchange gives comparable PPL and stronger source retrieval when Full layers provide the long-range path.

\subsection{Edge placement matters at equal budget}
\textbf{Would any equally sized repair work?} Random-Edge keeps the block prefix and adds $h^2$ edges per adjacent block pair through fixed random source/target permutations. It matches pure PBB's added-edge count, degree histogram, normalization, and parameters, but reaches PPL 25.63 versus 24.05 and probe accuracy 42.02\% versus 46.80\%. The location of the repair matters, not just the added access. Pure Block supplies the barrier check: exactly 25\% accuracy and source gain below $10^{-8}$ on 1,280 unreachable groups.

\textbf{Why not spend the same budget on wider blocks?} WideBlock-192 exactly matches PBB's 8K score count, including the final partial block, and both hybrids retain five Full layers. WideBlock reaches PPL 21.67 versus PBB's 21.27. Probe accuracy is close (80.57\% versus 81.31\%; paired difference interval $[-0.13,1.62]$ points), while source gain favors PBB by 0.207 nats, with interval $[0.162,0.252]$. Needle accuracy is also close (55.95\% versus 55.80\%). The distinction appears in Semantic exact retrieval: PBB improves accuracy from 54.05\% to 58.07\% and source gain from 1.332 to 1.975 (Appendix~\ref{app:retrieval_families}). At equal attention arithmetic, targeted repair combines competitive literal retrieval with better Semantic exact retrieval and language modeling.

\subsection{Natural-text evidence by position}
\begin{figure}[t]
 \centering\includegraphics[width=\linewidth]{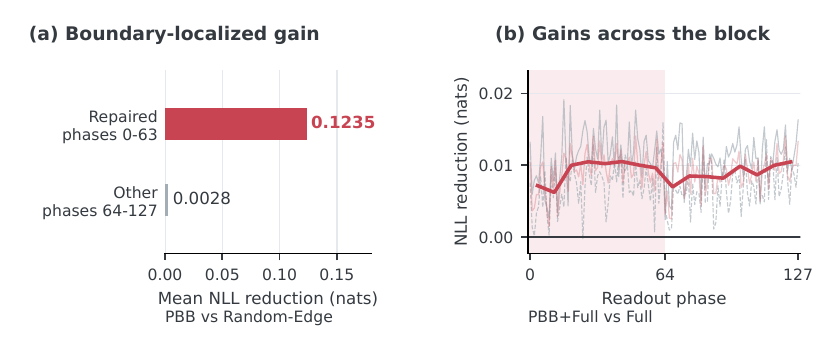}
 \caption{\textbf{Boundary-focused repair, benefits across the block.} Positive denotes lower PBB NLL. (a) Pure PBB versus Random-Edge, means over each half-block. (b) Full minus PBB+Full: gray curves are matched runs; red is their mean, with eight-phase averages in bold. Shading marks repaired phases.}
 \label{fig:phase}
\end{figure}
After excluding readouts with an end-of-document token in the preceding 128-position history, pure PBB reduces NLL relative to Random-Edge by 0.1235 nats over repaired phases, versus 0.0028 over the other half-block (Figure~\ref{fig:phase}). The boundary reduction is 1.503 nats; excluding phase zero still leaves a 0.1016-nat reduction over phases 1--63. The localized benefit therefore extends beyond the boundary spike. PBB+Full improves mean clean NLL over Full at every block phase; the per-run phase counts are reported in Appendix~\ref{app:recipe}.

On 128 matched held-out spans, shortening context from 8K to 128 tokens increases suffix NLL by 0.2002 for PBB+Full, 0.1999 for SWA+Full, and 0.2136 for Full. PBB+Full has lower absolute suffix loss at all four tested lengths, but its truncation penalty is similar to SWA+Full's: the paired difference is 0.0003 nats, with interval $[-0.0098,0.0103]$. This test scores identical suffixes outside the repaired half-block (Appendix~\ref{app:context}).

\section{Retrieval Gains Across Dense and MoE Models}
\label{sec:scale}
\begin{table}[t]
 \centering\small\begin{tabular}{llrrrr}
\toprule
Model & Architecture & PPL & Probe (\%) & Source gain & LM-6 (\%) \\
\midrule
350M & Full & \textbf{17.13} & 73.62 & 1.983 & 49.19 \\
350M & PBB+Full & 17.30 & \textbf{74.79} & \textbf{2.339} & \textbf{49.55} \\
2B MoE & Full & \textbf{14.25} & 75.62 & 2.326 & 52.02 \\
2B MoE & PBB+Full & 14.35 & \textbf{77.77} & \textbf{2.714} & \textbf{52.94} \\
\bottomrule
\end{tabular}

 \caption{\textbf{Stronger source retrieval with near-Full quality across model families.} Final checkpoints with matched exposure within each pair. MoE has 2.07B stored/300M active parameters. Probe results use the same 6,720-row protocol. Dense/MoE PPL uses 8M/1M held-out targets, respectively, matched within pairs.}
 \label{tab:scale}
\end{table}
\textbf{Does the design transfer beyond the 200M comparison?} A wider dense model and a sparsely activated MoE both retain near-Full PPL while improving source use (Table~\ref{tab:scale}). At 350M, PBB+Full improves probe accuracy, source gain, and LM-6 within 1.02\% of Full's PPL. Needle and Semantic exact retrieval gain 0.80 and 2.68 points, with higher source gain on both (Appendix~\ref{app:retrieval_families}).

\paragraph{Larger retrieval gains on the 2B MoE.}
After 6B training tokens, the MoE hybrid raises needle accuracy from 51.76\% to 56.88\% (+5.12 points) and Semantic exact retrieval from 52.44\% to 66.46\% (+14.02). Source gains rise from 1.347 to 1.683 and 2.061 to 2.917 nats (Appendix~\ref{app:retrieval_families}), pairing higher accuracy with stronger source sensitivity. The original probe improves from 75.62\% to 77.77\%, including 45.54\% versus 40.77\% over distances 2,048--6,144 (Appendix~\ref{app:distance_effects}).

\paragraph{Preserving modeling quality and improving downstream scores.}
The PPL gap to Full is 0.68\%. LM-6 gains 0.925 points, with task-stratified paired 95\% interval $[0.194,1.655]$ (Appendix~\ref{app:downstream_paired}). All six task point estimates, BLiMP, and ARC-Challenge favor the hybrid; ARC-Easy and ARC-Challenge gain 1.68 and 1.45 points. Stronger retrieval thus coexists with near-Full LM quality and improved downstream performance.

\section{Related Work}
Structured sparsity spans Sparse Transformers, Longformer, BigBird, and ETC \citep{child2019sparse,beltagy2020longformer,zaheer2020big,ainslie2020etc}; related local designs also appear in image generation \citep{parmar2018image}. Swin shifts partitions across layers \citep{liu2021swin}. LongLoRA uses shifted sparse attention for efficient fine-tuning with Full attention at inference \citep{chen2024longlora}. PBB instead preserves its boundary-repair pattern in training and inference, studying its role between periodic Full layers.

Hybrid H3 combines structured state spaces with selected attention layers \citep{fu2023h3}. Recurrence/compression \citep{dai2019transformerxl,hutchins2022block,rae2019compressive,munkhdalai2024infini} and content-dependent routing \citep{kitaev2020reformer,roy2021routing,mohtashami2023landmark,tang2024quest,jiang2024minference,yuan2025native,lu2025moba} offer other communication mechanisms. Differential Transformer changes attention weights to improve context use \citep{ye2025differential}; our intervention is static edge allocation. These are complementary design axes.

FlashAttention provides IO-aware exact execution \citep{dao2022flashattention,dao2024flashattention2}. FastGen adapts KV retention to profiled attention behavior \citep{ge2024fastgen}; PBB's ring instead follows its prescribed support exactly, without content-dependent eviction. Head/cache policies \citep{xiao2023streamingllm,xiao2024duoattention} and hierarchical layouts \citep{yu2023megabyte,ding2023longnet} offer further combinations. Long-context quality benchmarks \citep{hsieh2024ruler,bai2023longbench,modarressi2025nolima,liu2024lost} test capability rather than execution capacity.

\section{Conclusion}
The local graph should be designed for its role in the architecture. PBB preserves direct boundary exchange without retaining SWA's entire local relay range; matched comparisons, equal-arithmetic controls, and dense/MoE evaluations show stronger source retrieval with near-Full modeling quality. The same regular support enables exact ring storage and Flash-PBB's $1.82\times$ decode-core throughput over Flash-SWA, with half the allocated local KV and a smaller execution footprint. Together with whole-model training and inference gains, these results make targeted boundary exchange a practical design principle: preserve useful short paths, let global layers carry long-range communication, and exploit the resulting structure in the kernel.
\label{maintextend}

\clearpage
\section*{AI Use Statement}
This research was led by the author. AI tools assisted with theoretical and experimental development, data preparation, implementation and evaluation, result analysis, and manuscript preparation. The author made the final research decisions and takes full responsibility for the manuscript and associated artifacts.

\section*{Reproducibility Statement}
Appendices~\ref{app:proofs}--\ref{app:efficiency} specify equations, proofs, architecture dimensions, training recipes, evaluation construction, and timing protocols. Tables and figures retain source-file and protocol provenance. The preprint source package contains the manuscript, figures, tables, and bibliography; compilation requires neither model weights nor evaluation datasets. Numerical plotting inputs, a standalone PBB kernel with multi-block forward/backward checks, and the full training/evaluation artifact are maintained separately with the experiment records. Reproducing the exact filtered training stream additionally requires the retained-document selection; the screening protocol is described separately from the private lexicon.

\section*{Ethics Statement}
These research language models are not instruction-tuned assistants. Web-derived training text is screened for explicit sexual promotion while retaining ordinary biological and health-related material; screening does not guarantee absence of harmful content. No human-subject study is conducted. Public benchmark text is evaluated locally, and server performance tests use synthetic inputs. Model outputs can contain factual errors and should not be treated as authoritative.

{\raggedright\emergencystretch=2em
\bibliography{references}
\bibliographystyle{plainnat}
}
\clearpage
\appendix
\section{Proofs and Exact Arithmetic Budgets}
\label{app:proofs}
\subsection{Structural dependency sufficiency}
Fix a mask schedule, model parameters, and any auxiliary randomness shared between the compared forward passes. Input embeddings at position $t$ depend on that token and fixed positional information, so agreement on $\R_0(t)=\{t\}$ implies identical initial states. Suppose the statement holds at layer $\ell$. The residual input at $t$ and every key/value source in $\A_{\ell+1}(t)$ depend only on the sets included in Equation~\eqref{eq:closure}. Agreement on their union makes these vectors identical. Query/key projections, normalization, attention, residual addition, and all positionwise operations then produce identical states at $t$. This proves sufficiency by induction. Parameter values may suppress an allowed dependency; they cannot create one outside the structural closure.

\subsection{Fixed-block barrier and boundary-copy distribution}
For the fixed-block mask, every source $s\in\A(t)$ satisfies $p(t)\leq s\leq t$. Inductively, $\R_\ell(s)\subseteq\{p(t),\ldots,s\}$, and taking their union preserves the block-prefix bound. No depth introduces a source before $p(t)$.

To construct the task, choose a boundary $p>0$ and draw $Y$ uniformly from $K$ distinct candidate labels. Encode $Y$ only at position $p-1$. Draw the accessible block prefix and any fixed query independently of $Y$, and ask the readout at $p$ to predict $Y$ (or its corresponding next-token label). The model's readout is independent of $Y$ by the barrier. Its top-1 success probability is therefore at most $1/K$. For any output distribution $q$ on the candidate labels, with possible probability mass outside that set,
\begin{equation}
 \mathbb E[-\log q(Y)]
 =-\frac{1}{K}\sum_{k=1}^K\log q(k)
 \geq-\log\!\left(\frac{1}{K}\sum_{k=1}^K q(k)\right)
 \geq\log K.
\end{equation}
The first inequality is Jensen's inequality. The construction requires no redundant target-side cue or label-dependent accessible text. Natural retrieval tasks without this independence condition need not obey the same bound. Adding the direct causal edge $(p-1,p)$ removes this structural obstruction; it enables transmission, rather than guaranteeing that arbitrary trained parameters perform the task.

\subsection{Sliding-window and PBB relay}
For a sliding window, one layer expands an interval by at most $w-1$ positions to the left, with every intermediate source present. Induction yields Equation~\eqref{eq:swa_relay}, including the finite-sequence truncation at position zero. A Full layer reads position $s$ for every $s\leq t$, and each $\R_\ell(s)$ contains $s$ through its residual path. Its union is exactly $\mathcal C_t$. Later residual causal layers preserve this closure and cannot include future positions.

For the default $b=2h$, PBB can also be written using half-block chunks. Let $c(t)=\lfloor t/h\rfloor$. Then
\begin{equation}
 \A_{\PBB}(t)=\{\max(0,h(c(t)-1)),\ldots,t\},\qquad
 \R_L^{\PBB}(t)=\{\max(0,h(c(t)-L)),\ldots,t\}\quad(L\geq1).
 \label{eq:pbb_relay}
\end{equation}
For an even chunk, the preceding chunk is the previous block's repaired tail; for an odd chunk it is already in the same block. This proves the first equality. Each additional layer admits one more preceding half-block chunk, proving the second. PBB therefore also relays information across layers. At the default widths, its relay expands more slowly than SWA's; periodic Full layers remove the need to obtain global structural coverage from local relay alone.

\subsection{Direct coverage and single-softmax fusion}
Write a source as $s=t-d$ with nonnegative backward distance $d$. The block support gives $d\leq r$. The bridge-exclusive support gives $r<d\leq r+h$, and exists only for $p>0$ and $r<h$. These disjoint cases yield the coverage expression in Section~\ref{sec:method}. Since $r\leq2h-1$ and $r+h\leq2h-1$ for covered targets, every PBB edge satisfies $d<w=2h$ and is an SWA edge. The two patterns are therefore not coverage-incomparable in this setting.

For two disjoint supports $S_1,S_2$, define $Z_i=\sum_{s\in S_i}e^{z_s}$ and $o_i=Z_i^{-1}\sum_{s\in S_i}e^{z_s}v_s$. Then $(Z_1o_1+Z_2o_2)/(Z_1+Z_2)$ is exactly attention on $S_1\cup S_2$. Substituting $s_i=\log Z_i$ and dividing by $e^{\max(s_1,s_2)}$ gives the stable fusion
\begin{equation}
 o_t=\frac{e^{s_1-m}o_1+e^{s_2-m}o_2}{e^{s_1-m}+e^{s_2-m}},\qquad m=\max(s_1,s_2).
 \label{eq:lse}
\end{equation}
An empty bridge uses $Z_2=0$ and returns $o_1$. Overlapping supports must be deduplicated; otherwise shared sources receive duplicated exponential mass and the function changes.

\subsection{Exact arithmetic matching}
Let $E$ denote the number of unique causal scores per query head. For $n\geq w$ and $n$ divisible by $b$,
\begin{align}
 E_{\Full}&=n(n+1)/2, &
 E_{\rm Block}&=n(b+1)/2,\\
 E_{\SWA}&=nw-w(w-1)/2, &
 E_{\PBB}&=n(b+1)/2+(n/b-1)h^2.
\end{align}
For an arbitrary block size $a$, set $q=\lfloor n/a\rfloor$ and $u=n-qa$. Then
\begin{equation}
 E_{\rm Block(a)}=qa(a+1)/2+u(u+1)/2.
\end{equation}
At $n=8192$, $a=192$ gives $q=42$, $u=128$, and $E=786432$, exactly equal to PBB with $b=128,h=64$. Random-Edge selects 64 target phases and 64 previous-block source phases per boundary using seeded bijections; it therefore adds exactly $64^2$ edges and matches PBB's added-degree histogram. Its fixed selection neither depends on model activations nor resamples during training.

\begin{table}[htbp]
 \centering\small
 \begin{tabular}{lrr}
 \toprule
 Local pattern & Scores/head/local layer & Scores/head/20-layer 3:1 hybrid\\
 \midrule
 Full &33,558,528&671,170,560\\
 Block-128 &528,384&175,718,400\\
 SWA-128 &1,040,448&183,399,360\\
 PBB-128/64 &786,432&179,589,120\\
 Random-Edge &786,432&179,589,120\\
 WideBlock-192 &786,432&179,589,120\\
 \bottomrule
 \end{tabular}
 \caption{Exact mathematical score counts at $n=8192$. The Full row is all-Full. The Random-Edge hybrid count is an arithmetic reference; the trained Random-Edge control is pure local.}
\end{table}
Computing both $QK^\top$ and $AV$ requires approximately $4d_hE$ multiply/add FLOPs per query head under the usual two-FLOP multiply-accumulate convention. Matching $E$, dimensions, projections, and other modules therefore matches the mathematical attention arithmetic for the designated controls. Kernel padding, tile layout, memory traffic, and launch overhead can still differ. Full, SWA, and PBB do \emph{not} have identical model FLOPs simply because their token budgets match.

\section{The Value of Local Exchange with Global Communication}
\label{app:local_value}
The design hypothesis is to retain useful local information exchange rather than maximize the range of local relay once a global mechanism is available. Three quantities must be distinguished: whether a path exists, how soon an intermediate feature can be transmitted, and how strongly that transmission affects a prediction. The first two admit graph statements. The third depends on learned transformations and the task. The following analysis connects these quantities without identifying them.

\subsection{Input coverage versus intermediate-feature latency}
Use a layered graph with nodes $(\ell,t)$, denoting the representation at position $t$ after layer $\ell$. An allowed attention or residual edge connects $(\ell,s)$ to $(\ell+1,t)$. Define
\begin{equation}
 \tau_\ell(s,t)=\min\{a\geq 1:(\ell,s)\leadsto(\ell+a,t)\},\qquad s<t,
 \label{eq:communication_latency}
\end{equation}
with $\tau_\ell(s,t)=\infty$ if there is no such path. This is the minimum number of additional layers required for possible transmission. It is not the number of non-self attention hops on a fixed-depth path: residual edges can preserve a feature while it waits for a later cross-position edge.

\begin{proposition}[Boundary exchange between Full layers]
Assume causal attention is the only cross-position operation, residual self paths are present, and each $k$th layer is Full. Let $f>\ell$ be the next Full layer, which is assumed to exist. For a fixed-block boundary $p>0$, choose valid sequence positions $s\in\{p-h,\ldots,p-1\}$ and $t\in\{p,\ldots,p+h-1\}$, with $b=2h$. Then
\begin{equation}
 \tau_\ell^{\mathrm{Block+Full}}(s,t)=f-\ell\leq k,
 \qquad
 \tau_\ell^{\mathrm{PBB+Full}}(s,t)
 =\tau_\ell^{\mathrm{SWA+Full}}(s,t)=1,
 \label{eq:boundary_latency}
\end{equation}
where SWA uses $w=b$. For a source and target within one block, Block+Full also permits next-layer exchange.
\label{prop:boundary_latency}
\end{proposition}
\paragraph{Proof.}
Before layer $f$, fixed-block attention and residual edges cannot leave the source block. A residual path preserves access to position $s$ until layer $f$, whose Full mask contains the edge to $t$. Hence the first possible cross-block arrival is $f-\ell$. For the specified source and target ranges, PBB contains $(s,t)$ in its next local layer; the edge is also present if that layer is Full. Moreover $t-s\leq2h-1=w-1$, so the same conclusion holds for SWA. Within a block, all three local masks contain the causal source--target edge. This proves the claims.

The proposition concerns a fresh perturbation at $(\ell,s)$ with all other layer-$\ell$ states held fixed. Even if an earlier Full layer made every input token reachable, that layer has not broadcast this newly introduced intermediate perturbation. Thus input-level connectivity can be saturated while communication latency remains different. For a 3:1 schedule, a perturbation introduced just after a Full layer can cross the specified boundary in one additional PBB layer, instead of waiting four layers under Block+Full. Pure fixed-block attention has infinite cross-block latency; it is not merely a slower relay.

PBB matches SWA's one-layer exchange for these boundary pairs with 24.4\% fewer total local scores at the default 8K geometry. The proposition therefore identifies a latency-preserving allocation: retain direct boundary exchange while leaving global communication to Full layers. Its prediction-level benefit depends on the learned transformations and task.

\subsection{Weighted paths and effective influence}
At a differentiable forward pass, write the block Jacobian as
\begin{equation}
 J_\ell(t,s)=\frac{\partial H_t^{(\ell)}}{\partial H_s^{(\ell-1)}},
 \qquad B_\ell(t,s)\geq\|J_\ell(t,s)\|_{\mathrm{op}}.
\end{equation}
Choose the nonnegative matrix $B_\ell$ to be zero outside the attention and residual support. The chain rule and triangle inequality give
\begin{equation}
 \left\|\frac{\partial H_t^{(L)}}{\partial H_s^{(0)}}\right\|_{\mathrm{op}}
 \leq (B_LB_{L-1}\cdots B_1)_{ts}
 =\sum_{\pi:s\leadsto t}\prod_{\ell=1}^{L}B_\ell(\pi_\ell,\pi_{\ell-1}).
 \label{eq:path_influence}
\end{equation}
Structural reachability determines where such paths are allowed, not their weights. These Jacobians include query/key dependence, normalization, residuals, and positionwise transformations; attention probabilities alone are not the full influence operator. For MoE, a local Jacobian description requires a neighborhood without a routing switch, or a finite-perturbation analysis instead.

Even a nonnegative normalized linear attention model does not imply exponential attenuation with hop count. Let $S$ shift a scalar signal by one causal position and $T=(I+S)/2$. Away from the sequence boundary, a source $m$ positions earlier contributes to $T^{2m}$ with coefficient
\begin{equation}
 (T^{2m})_{t,t-m}=\binom{2m}{m}4^{-m}
 \sim\frac{1}{\sqrt{\pi m}}.
 \label{eq:relay_counterexample}
\end{equation}
Each complete path has weight $4^{-m}$, but there are $\binom{2m}{m}$ paths. Bounding one path does not bound their sum by the same factor. The deterministic shift $T=S$ preserves influence exactly along arbitrarily many hops; sharply peaked softmax weights can approximate this behavior at any fixed depth. General neural transformations can also amplify or reconstruct signals. An exponential influence law therefore needs additional contraction and path-multiplicity assumptions, which the present checkpoint evaluations do not establish.

Equation~\eqref{eq:path_influence} nevertheless supplies a measurement target: compare influence by layer distance, source distance, and boundary phase. Effective receptive field should be defined using a declared influence or task metric and threshold, rather than substituted for the structural set $\R_L(t)$. Greater structural range permits more dependencies but need not increase any particular source's measured effect.

\subsection{A sufficient condition for inexpensive edge removal}
For a single head, hold query, keys, and values fixed. Let $A$ be a reference support, $S\subseteq A$ a nonempty retained support, and $D=A\setminus S$. Write $a_s$ for the softmax probabilities on $A$, and $\mu=\sum_{s\in D}a_s$. Let $o_A,o_S,o_D$ be the outputs normalized on their respective supports. For nonempty $D$,
\begin{equation}
 o_A=(1-\mu)o_S+\mu o_D,\qquad
 \|o_A-o_S\|=\mu\|o_D-o_S\|\leq 2V\mu,
 \label{eq:omitted_mass}
\end{equation}
where $V=\max_{s\in A}\|v_s\|$. If $D$ is empty the difference is zero. The identity follows by splitting the numerator of attention into retained and removed terms. Small removed probability mass is sufficient for a small output change; it is not necessary, since retained and removed value averages may already be similar. Conversely, low edge count alone does not ensure a small change. Output projection multiplies the bound by its operator norm.

This identity applies to an SWA-to-PBB support restriction at the same hidden state, because default PBB is a subset of SWA. It makes the source-competition hypothesis concrete: removing sources renormalizes retained weights, while its immediate effect depends on both removed mass and value contrast. It does not imply that removed sources are noise or that renormalization improves prediction.

To extend the calculation across layers, let $F_\ell$ and $\widetilde F_\ell$ be the reference and restricted layer maps. Suppose the restricted map has Lipschitz constant $K_\ell$ on the relevant states and the same-state map discrepancy is bounded by $\epsilon_\ell$. From identical initial states, the state error obeys
\begin{equation}
 \Delta_\ell\leq K_\ell\Delta_{\ell-1}+\epsilon_\ell,
 \qquad
 \Delta_L\leq\sum_{\ell=1}^{L}\epsilon_\ell\prod_{j=\ell+1}^{L}K_j.
 \label{eq:restriction_error}
\end{equation}
This follows by adding and subtracting $\widetilde F_\ell$ evaluated at the reference state, then applying the Lipschitz bound and induction. Neither small $\epsilon_\ell$ nor nonamplifying $K_\ell$ follows from the mask geometry. The separately trained PBB and SWA models also have different weights; Equation~\eqref{eq:restriction_error} is a same-checkpoint intervention model, not a derivation of their PPL ordering.

\subsection{Relation to DeepSeek-V4.1-Flash}
\label{app:deepseek_replay}
\citet{deepseekai2026v41flash} provide a production-scale example of prioritizing recent local exchange while retaining global history. Their Causal Encoder-Decoder has 20 encoder and 20 decoder layers. Section 2.2 gives an exact decoder SWA reconstruction span of an additional $n_{\rm win}L/2$ prompt tokens, where $L/2$ denotes decoder depth. Citing effective receptive fields smaller than theoretical range \citep{chen2025powerattention}, the report instead replays the most recent $n_{\rm win}$ tokens. Section 3.2.2 regenerates local SWA states while reusing cached global KV information; decoder replay uses recent encoder outputs and globally projected encoder KV. The report describes little quality loss from this approximate reconstruction and also simulates decoder replay during post-training.

This observation motivates our allocation hypothesis: when a global path retains long-range information, useful nearby transfer may justify more local computation than exact restoration of the full relay history. The interventions are distinct: bounded replay changes cache reconstruction with global sparse retrieval retained, whereas PBB changes local attention edges in a periodic-Full model. DeepSeek's CSA2 ``Full Mode'' differs from our all-prefix Full layer, and runtime local caches remain in its design. We use the reported quality observation as motivation, not as a PBB evaluation or influence-decay theorem; the report's conclusion also identifies approximate reconstruction and sparse retrieval as possible error sources in untested cases.

The engineering connection is the separate treatment of local and global state. DeepSeek removes SWA state from long-lived persistent caching, keeps a short-lived local-state pool, and uses bounded replay on misses. PBB instead gives each local layer a phase-aligned support whose runtime KV can be maintained without relocation, while Full layers retain global history. Both designs avoid imposing the global state's storage or reconstruction policy on local state. Their mechanisms and memory scopes differ: bounded replay approximates cross-layer reconstruction, whereas Flash-PBB's ring layout exactly implements the prescribed local support. Section~\ref{sec:efficiency} supplies independent PBB measurements for this topology-to-execution connection.

\subsection{Empirical implications and future interventions}
Three observations support selective local exchange. Random-Edge controls show that edge placement matters at equal budget. Natural-text loss differences between pure PBB and Random-Edge concentrate at the repaired phases. The pure-local versus hybrid comparison shows that a smaller local support can yield higher source-retrieval accuracy when Full layers are present. In the context ablation, PBB+Full maintains lower absolute suffix NLL across all four lengths while the two hybrids have similar truncation penalties (Appendix~\ref{app:context}). These results connect targeted boundary exchange to measured behavior; the path-influence mechanism remains a hypothesis.

Future same-checkpoint interventions can separate these mechanisms. A source-dependent perturbation introduced at one layer-position node, with other states held fixed, would measure the arrival latency in Equation~\eqref{eq:boundary_latency} and its final prediction effect. Removing bridge-exclusive edges versus matched-budget edges elsewhere would relate removed attention mass and value contrast to phase NLL and source gain. A global-KV-preserving replay intervention would separately test sensitivity to local reconstruction history; ordinary input truncation removes history from the global path too.

\section{Architecture, Data, and Optimization}
\label{app:recipe}
\begin{table}[htbp]
 \centering\small
 \begin{tabular}{lrrr}
 \toprule
 Setting &200M dense&350M dense&2B MoE\\
 \midrule
 Stored parameters &205,417,600&351,807,872&2,069,563,136\\
 Active parameters &205,417,600&351,807,872&300,091,136\\
 Layers &20&20&20\\
 Hidden width &640&896&768\\
 Dense FFN width &2,304&3,328&--\\
 Query/KV heads &10/2&14/2&12/2\\
 Head dimension &64&64&64\\
 Routed experts / top-$k$ &--&--&64/4\\
 Routed / shared FFN width &--&--&640/800\\
 Padded vocabulary &151,936&151,936&151,936\\
 Sequence length &8,192&8,192&8,192\\
 Tokens/update &393,216&393,216&393,216\\
 Updates &10,449&18,000&15,259\\
 Training tokens &4,108,713,984&7,077,888,000&6,000,082,944\\
 \bottomrule
 \end{tabular}
 \caption{Architecture and exposure. Dense parameter counts include tied token embeddings. MoE active and stored counts are reported separately. Sparse/full comparisons keep these dimensions fixed.}
\end{table}

All models use RoPE with base $10^6$, RMSNorm with $\epsilon=10^{-6}$, tied input/output embeddings, zero dropout, and initialization standard deviation 0.02. Attention uses BF16 computation with FP32 training master parameters. The common AdamW recipe \citep{loshchilov2019decoupled} uses $\beta_1=0.9$, $\beta_2=0.95$, $\epsilon=10^{-8}$, weight decay 0.1, and gradient clipping at norm 1.0. The maximum/minimum LR is $3\times10^{-4}/3\times10^{-5}$. The schedule is warmup--stable--decay: a 5.56\% linear warmup, a plateau to 88.89\%, and cosine decay over the remaining updates. Integer step boundaries are recorded in each run configuration; for the 18,000-step run they are 1,000 and 16,000. For 200M they are 580 and 9,288. Physical micro-batch and gradient accumulation can vary with memory constraints while their product remains 48 sequences. This is a common exposure recipe, not a separate per-architecture learning-rate search.

\paragraph{Training runs and result provenance.}
Full/PBB comparisons use two paired training seeds, 13 and 2025, at 200M, and one training seed, 13, for each architecture at 350M and 2B MoE. Other 200M controls use seed 13. In Table~\ref{tab:seeds}, run A denotes seed 13 and run B denotes seed 2025; slash-separated 200M effects in the main text follow this order. The local-family and scaling tables report seed 13. Figure~\ref{fig:overview} uses the seed-13 MoE trajectories. In Figure~\ref{fig:phase}b, the gray curves are the two paired 200M runs and the red curve is their mean: PBB+Full improves clean NLL at 128/128 phases for seed 13 and 127/128 for seed 2025. Sample-bootstrap intervals describe evaluation-sample uncertainty, not variation across training seeds.

The main dense comparisons train on the same filtered token stream; seeds control initialization and sampling state. The MoE comparison likewise shares its data exposure and model dimensions. Scaling the dense family changes attention width and FFN width together, not only the FFN. The MoE experiment changes feed-forward organization and total capacity as well, so it tests transfer of the attention design across model families.

MoE top-4 probabilities are renormalized over the selected experts. The router load-balancing auxiliary-loss coefficient is $10^{-3}$; reported validation/test NLL excludes this auxiliary term. All 20 feed-forward layers use routed and shared experts, with Qwen2-MoE implementation semantics, rather than substituting a different attention or recurrent module. The active-parameter count includes common parameters, router parameters, shared-expert parameters, and $4/64$ of the routed-expert parameters. The MoE LR boundaries are 848 warmup updates and decay starting at update 13,563. Activation checkpointing is enabled in training.

\paragraph{Data preparation.}
FineWeb-Edu documents undergo a conservative screening stage for explicit sexual promotion, followed by additional keyword screening and sampled manual review. Ordinary biological and health-related educational material is retained when not excluded by the conservative rules. Thirteen manually identified documents are excluded from the retained corpus; clean training text backfills their 24,594 tokens. Screening is imperfect and is not a guarantee of content safety. The model and evaluation implementation is separable from the private screening lexicon; reproducing the exact filtered corpus requires the retained-document selection, not just an informal statement that filtering occurred.

Document hashes determine disjoint train, validation, and test assignment before token packing. The retained training stream has 15,728,640,001 tokens. Validation and test each have 8,388,608 tokens; the documented test split contains 7,876 documents. The Qwen2.5 tokenizer and its end-of-text token 151643 are used throughout the main experiments. Training runs consume prefixes according to their budgets. Test data are never used for gradient updates.

\paragraph{PPL and phase conventions.}
Held-out PPL is $\exp(\text{mean token NLL})$, not the arithmetic mean of per-sequence perplexities. Dense models score 8,388,608 test targets; MoE pairs score the same 1,048,576-target prefix. Thus each architecture comparison is token matched, while cross-family absolute PPLs also differ in evaluation sample size. The readout phase is the position of the hidden state predicting the next token; the predicted-token phase is $(r+1)\bmod b$. The clean-NLL statistic excludes readouts with an end-of-text token in the 128-position input history ending at that readout. Phase comparisons use identical counts and the unsmoothed per-phase values; plotting averages do not change the reported statistics.

\section{Formal Structural and Likelihood Evaluation}
\label{app:probes}
\subsection{Balanced source rotation}
The formal v4 probe freezes 16 background/template anchors per distance--phase cell, with 15 distances and seven readout phases. Each cell has its own background spans, giving 1,680 distinct corpus offsets. The distance grid is
\[
 \{8,16,32,64,65,96,127,128,160,256,512,1024,2048,4096,6144\}.
\]
Readout phases are $\{0,16,32,63,64,96,127\}$. The four candidate strings are \texttt{ red}, \texttt{ blue}, \texttt{ green}, and \texttt{ yellow}, each a single token, with IDs 2518, 6303, 6176, and 13753. Natural held-out text provides the background. Four templates state a secret word, recorded answer, key value, or code word. Source and readout positions are stored explicitly, with theoretical distance defined as their token-index difference.

Each group fixes background, positions, and template, then rotates all four source labels once. There are $16\times15\times7=1680$ groups and 6,720 rows. A candidate-prior-only predictor cannot improve over 25\% within a complete group. Let $x^{(j)}$ be the prompt with source label $j$, and $v_j$ its target token. Define
\begin{equation}
 G=\frac14\sum_{j=1}^4\left[
 \log p(v_j\mid x^{(j)})-\frac13\sum_{k\ne j}\log p(v_j\mid x^{(k)})
 \right].
\end{equation}
The probabilities are normalized over the full model vocabulary. Positive $G$ measures increased probability for a label when the source supports it, beyond the same label's probability under counterfactual sources. Accuracy selects the most likely candidate. Each group is evaluated as a unit for invariance and paired analyses; its four rows are not four independent background samples.

In a complete balanced group, the vocabulary log-partition terms cancel. Writing $a_j^{(k)}$ for the logit of label $j$ under source rotation $k$ gives the equivalent statistic
\begin{equation}
 G=\frac14\sum_{j=1}^4\left[a_j^{(j)}-\frac13\sum_{k\ne j}a_j^{(k)}\right].
\end{equation}
Each prompt's log-partition appears once with coefficient $-1/4$ and three times with coefficient $+1/12$. Thus $G$ is a counterfactual source-sensitive logit contrast, not a measure of full-vocabulary calibration. The recorded log-probability evaluation and this identity compute the same balanced-group quantity.

The manifest was frozen after Full-only calibration and before comparative architecture evaluation. WideBlock uses a 192-phase metadata view of the same frozen token tensors; its source positions and candidate labels do not change. Theoretical nonreachability requires identical logits across source rotations; the evaluator checks this with absolute tolerance $10^{-5}$.

\paragraph{Numerical repeatability.}
The final 350M Full result uses all 6,720 rows at step 18,000: 73.6161\% accuracy and source gain 1.983262. A separate complete evaluation gives 73.4970\% and 1.983149. This small BF16 backend-sensitive variation preserves the Full/PBB ordering. A 32-row replay with FlashAttention 2.8.3.post1 before resumption showed maximum candidate-logit deviation 0.0625 and no candidate-ranking disagreement.

\subsection{Paired uncertainty}
For each pair, group IDs, source positions, and backgrounds are matched. We resample the 16 complete source-rotation groups within each of the 105 distance--phase cells, preserving equal cell weights, with 10,000 replicates and RNG seed 20260905 \citep{koehn2004statistical,berg2012empirical,dror2018hitchhiker}. Table~\ref{tab:paired} reports percentile 95\% intervals with each complete four-label group as the resampling unit. Bootstrap intervals throughout this appendix describe sample uncertainty conditional on the evaluated checkpoints; training-seed replication is reported separately in Table~\ref{tab:seeds}. The following sensitivity analysis examines overlapping backgrounds.

\paragraph{Background-overlap sensitivity.}
A metadata audit finds 1,490 overlapping causal-prefix pairs among the 1,680 groups, forming 681 connected overlap components (largest: 15 groups). Using the full stored 8K spans gives 327 components (largest: 25). We additionally draw mean-one exponential weights per component, share them across its groups, normalize within each geometry cell, and average cells equally (10,000 draws; RNG seed 20260912). The resulting 2.5--97.5 percentile ranges retain positive accuracy differences for the five comparisons against Full or SWA and include zero for WideBlock; all six source-gain differences remain positive. For 350M accuracy, the causal-prefix/full-span ranges are $[0.20,2.14]$/$[0.22,2.12]$ percentage points; for MoE they are $[1.20,3.11]$/$[1.20,3.13]$. The source package includes all ranges and overlap counts. These are geometric-overlap sensitivity ranges, rather than document-cluster confidence intervals, and are reported alongside Table~\ref{tab:paired}.

\begin{table}[htbp]
 \centering\scriptsize\setlength{\tabcolsep}{3pt}\begin{tabular}{lrr}
\toprule
Paired comparison & Accuracy difference (pp) & Source gain (nats) \\
\midrule
200M, 13: PBB+Full $-$ Full & 11.47 [10.46, 12.50] & 0.898 [0.858, 0.938] \\
200M, 2025: PBB+Full $-$ Full & 11.82 [10.74, 12.95] & 0.594 [0.555, 0.633] \\
200M, 13: PBB+Full $-$ SWA+Full & 4.81 [3.96, 5.62] & 0.238 [0.203, 0.273] \\
200M, 13: PBB+Full $-$ WideBlock+Full & 0.74 [-0.13, 1.62] & 0.207 [0.162, 0.252] \\
350M, 13: PBB+Full $-$ Full & 1.18 [0.21, 2.14] & 0.356 [0.315, 0.397] \\
MoE, 13: PBB+Full $-$ Full & 2.14 [1.13, 3.15] & 0.388 [0.333, 0.442] \\
\bottomrule
\end{tabular}

 \caption{Paired PBB+Full differences with cell-stratified group-bootstrap intervals. Positive values favor PBB+Full. The small accuracy difference against WideBlock includes zero, while its source-gain interval is positive.}
 \label{tab:paired}
\end{table}

\subsection{Likelihood benchmarks}
\label{app:downstream}
All likelihood tasks use zero-shot evaluation without a chat template. LAMBADA uses the \texttt{lambada\_\allowbreak openai} task; HellaSwag, PIQA, ARC-Easy, SciQ, and ARC-Challenge report the harness's length-normalized accuracy. Winogrande and BLiMP report accuracy. LM-6 averages its six tasks equally. Table~\ref{tab:downstream} gives every recorded final-checkpoint task value; Table~\ref{tab:allfinal} separates validation and held-out PPL, and Table~\ref{tab:downstream_paired} reports paired task differences and intervals.

\begin{table}[t]
 \centering\scriptsize\setlength{\tabcolsep}{2.2pt}
 \begin{tabular}{lllrrrrrrrr}
\toprule
Model & Attention & Seed & LAMB. & Hella. & PIQA & ARC-E & SciQ & Wino. & BLiMP & ARC-C \\
\midrule
200M & Full & 13 & 23.13 & 30.52 & 60.83 & 44.49 & 63.40 & 52.33 & 78.06 & 25.09 \\
200M & Full & 2025 & 23.37 & 30.67 & 61.59 & 45.33 & 66.10 & 52.25 & 78.44 & 24.91 \\
200M & Block & 13 & 18.79 & 30.55 & 61.86 & 46.89 & 60.10 & 50.91 & 79.71 & 25.94 \\
200M & PBB & 13 & 22.47 & 30.45 & 61.10 & 45.92 & 60.10 & 49.64 & 80.37 & 25.60 \\
200M & SWA & 13 & 23.33 & 31.11 & 61.26 & 45.58 & 63.90 & 51.62 & 79.64 & 26.02 \\
200M & PBB+Full & 13 & 24.41 & 30.57 & 61.59 & 45.33 & 63.90 & 50.99 & 79.83 & 25.77 \\
200M & PBB+Full & 2025 & 23.37 & 30.94 & 61.43 & 45.12 & 64.70 & 50.36 & 79.90 & 25.26 \\
200M & SWA+Full & 13 & 24.90 & 30.69 & 60.83 & 45.20 & 63.00 & 50.28 & 79.89 & 26.02 \\
200M & WideBlock-192+Full & 13 & 22.20 & 31.10 & 62.02 & 44.07 & 64.80 & 51.30 & 80.37 & 25.43 \\
200M & Random-Edge & 13 & 21.11 & 30.50 & 61.48 & 44.87 & 62.20 & 51.46 & 79.74 & 25.94 \\
350M & Full & 13 & 28.16 & 35.79 & 64.36 & 49.45 & 67.10 & 50.28 & 80.75 & 25.34 \\
350M & PBB+Full & 13 & 28.45 & 35.67 & 64.09 & 49.03 & 69.60 & 50.43 & 80.40 & 25.60 \\
MoE & Full & 13 & 32.51 & 39.83 & 66.43 & 52.53 & 69.60 & 51.22 & 81.03 & 28.16 \\
MoE & PBB+Full & 13 & 32.84 & 39.99 & 67.30 & 54.21 & 71.00 & 52.33 & 81.09 & 29.61 \\
\bottomrule
\end{tabular}

 \caption{Final-checkpoint zero-shot likelihood results (percent). Hybrids use 3:1 local/full layers. ARC-E/ARC-C are Easy/Challenge. The eight task columns retain their specified metric definitions; LM-6 excludes BLiMP and ARC-C.}
 \label{tab:downstream}
\end{table}

\paragraph{Paired downstream analysis.}
\label{app:downstream_paired}
We analyze retained per-item scores for Full and PBB+Full at both 200M seeds and the MoE endpoint. All 269,544 item pairs match in task, document ID, document content, prompt arguments, target, and filter; recomputed accuracies reproduce the reported values. Table~\ref{tab:downstream_paired} gives all eight task differences. We draw 20,000 paired bootstrap replicates with RNG seed 20260912, resampling the same items for both models. BLiMP resamples sentence pairs within each of its 67 fixed subtasks and averages the subtask means; LM-6 independently resamples items within each of its six fixed tasks and averages tasks equally. We report pointwise percentile intervals.

BLiMP's gains at 200M remain positive in both seeds. As a separate sensitivity to subtask composition, resampling the 67 paired subtask means gives ranges $[0.68,3.04]$ and $[0.03,3.02]$ points; this does not establish independence among linguistic templates. ARC-Challenge's positive point estimates have intervals crossing zero in all three pairs. For MoE, the equally weighted LM-6 improvement has interval $[0.194,1.655]$, although most individual-task intervals cross zero. Retaining the same items jointly across the two fixed 200M seeds gives a mean LM-6 difference of $-0.109$ points, with interval $[-0.662,0.435]$.

The supplementary numerical files include discordant item counts, two-sided exact McNemar tests, and Holm adjustments over eight tasks within each pair and all 24 task comparisons. These tests assume independent item pairs; the BLiMP subtask sensitivity separately probes template dependence. Pointwise intervals are not multiplicity-adjusted: MoE ARC-Easy's interval excludes zero, but its within-pair Holm-adjusted $p$-value is 0.309. This secondary analysis adds uncertainty estimates without changing checkpoints, metrics, or the original point estimates.

\begin{table}[htbp]
 \centering\scriptsize\setlength{\tabcolsep}{3pt}
 \begin{tabular}{@{}lrrrrrr@{}}
\toprule
& \multicolumn{2}{c}{200M, seed 13} & \multicolumn{2}{c}{200M, seed 2025} & \multicolumn{2}{c}{2B MoE, seed 13} \\
\cmidrule(lr){2-3}\cmidrule(lr){4-5}\cmidrule(l){6-7}
Task & $\Delta$ & 95\% interval & $\Delta$ & 95\% interval & $\Delta$ & 95\% interval \\
\midrule
LAMBADA & $+1.281$ & $[+0.310,+2.251]$ & $+0.000$ & $[-0.912,+0.931]$ & $+0.330$ & $[-0.699,+1.358]$ \\
HellaSwag & $+0.050$ & $[-0.498,+0.597]$ & $+0.269$ & $[-0.279,+0.807]$ & $+0.159$ & $[-0.408,+0.747]$ \\
PIQA & $+0.762$ & $[-0.762,+2.285]$ & $-0.163$ & $[-1.687,+1.360]$ & $+0.871$ & $[-0.653,+2.394]$ \\
ARC-Easy & $+0.842$ & $[-0.715,+2.399]$ & $-0.210$ & $[-1.684,+1.263]$ & $+1.684$ & $[+0.126,+3.241]$ \\
SciQ & $+0.500$ & $[-1.700,+2.700]$ & $-1.400$ & $[-3.600,+0.800]$ & $+1.400$ & $[-0.500,+3.300]$ \\
Winogrande & $-1.342$ & $[-4.578,+1.896]$ & $-1.894$ & $[-5.209,+1.342]$ & $+1.105$ & $[-1.973,+4.183]$ \\
ARC-Challenge & $+0.683$ & $[-1.195,+2.560]$ & $+0.341$ & $[-1.536,+2.133]$ & $+1.451$ & $[-0.512,+3.413]$ \\
BLiMP & $+1.775$ & $[+1.488,+2.066]$ & $+1.458$ & $[+1.173,+1.743]$ & $+0.058$ & $[-0.207,+0.331]$ \\
\midrule
LM-6 & $+0.349$ & $[-0.431,+1.121]$ & $-0.567$ & $[-1.345,+0.200]$ & $+0.925$ & $[+0.194,+1.655]$ \\
\bottomrule
\end{tabular}

 \caption{Paired downstream differences (PBB+Full minus Full, percentage points) and pointwise 95\% bootstrap intervals. All comparisons use final checkpoints. LM-6 keeps six equally weighted tasks; BLiMP keeps 67 fixed subtasks. Full task values appear in Table~\ref{tab:downstream}.}
 \label{tab:downstream_paired}
\end{table}

\begin{table}[t]
 \centering\footnotesize\setlength{\tabcolsep}{3pt}
 \begin{tabular}{lllrrrr}
\toprule
Model & Attention & Seed & Val PPL & Test PPL & Probe (\%) & Source gain \\
\midrule
200M & Full & 13 & 22.26 & 21.52 & 69.84 & 1.645 \\
200M & Full & 2025 & 22.12 & 21.38 & 68.26 & 1.811 \\
200M & Block & 13 & 31.28 & 30.42 & 35.83 & 0.315 \\
200M & PBB & 13 & 24.82 & 24.05 & 46.80 & 0.647 \\
200M & SWA & 13 & 24.09 & 23.31 & 54.27 & 0.911 \\
200M & PBB+Full & 13 & 21.95 & 21.27 & 81.31 & 2.543 \\
200M & PBB+Full & 2025 & 21.94 & 21.22 & 80.07 & 2.405 \\
200M & SWA+Full & 13 & 22.00 & 21.29 & 76.50 & 2.305 \\
200M & WideBlock-192+Full & 13 & 22.40 & 21.67 & 80.57 & 2.336 \\
200M & Random-Edge & 13 & 26.43 & 25.63 & 42.02 & 0.576 \\
350M & Full & 13 & 17.80 & 17.13 & 73.62 & 1.983 \\
350M & Block & 13 & 25.94 & 25.04 & 35.19 & 0.387 \\
350M & PBB+Full & 13 & 17.97 & 17.30 & 74.79 & 2.339 \\
MoE & Full & 13 & 15.61 & 14.25 & 75.62 & 2.326 \\
MoE & PBB+Full & 13 & 15.69 & 14.35 & 77.77 & 2.714 \\
\bottomrule
\end{tabular}

 \caption{All final checkpoints used in the current dense/MoE record. Validation and test PPL use separate splits. Structural results use 6,720 balanced rows; blank downstream cells denote unavailable measurements.}
 \label{tab:allfinal}
\end{table}

\subsection{Local-family comparison}
\begin{figure}[t]
 \centering\includegraphics[width=\linewidth]{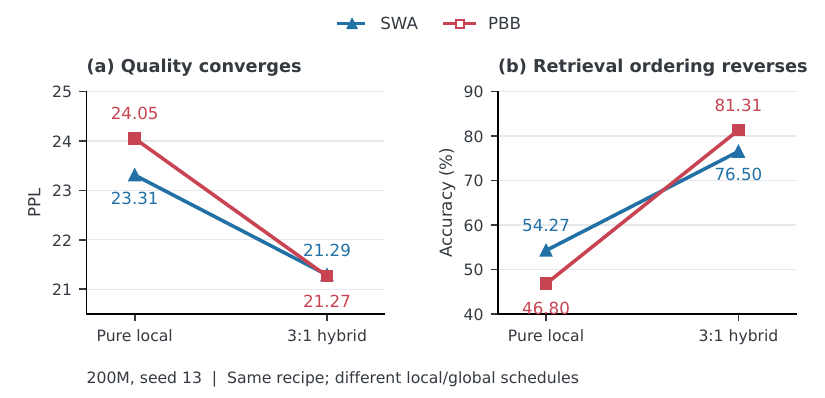}
 \caption{\textbf{Different local choices for different schedules.} At 200M, seed 13, pure SWA outperforms pure PBB; hybrid PPLs converge while source-retrieval ordering reverses. The four models share a training recipe; Appendix~\ref{app:proofs} gives their attention score budgets.}
 \label{fig:ordering}
\end{figure}

Figure~\ref{fig:ordering} shows the seed-13 SWA/PBB comparison alongside the paired Full/PBB results in Table~\ref{tab:seeds}. Pure SWA has lower PPL and higher structural accuracy than pure PBB. Under the shared periodic-full schedule, the hybrids have comparable PPL, while PBB+Full has higher structural accuracy. Thus the local pattern's role in the full architecture matters: the best pure-local choice need not be the best hybrid component.

\subsection{Source retrieval throughout training}
\label{app:probe_checkpoints}
Table~\ref{tab:probe_checkpoints} reports 5k, 10k, and final evaluations on the same frozen 6,720-row, four-candidate probe. PBB+Full exceeds Full at every listed checkpoint in both seeds and exceeds SWA+Full at all three seed-13 checkpoints. The endpoint advantage therefore also appears at intermediate stages within each training trajectory. Main comparisons use matched final training budgets.

\begin{table}[htbp]
 \centering\small\begin{tabular}{rrrrr}
\toprule
Seed & Step & Full & SWA+Full & PBB+Full \\
\midrule
13 & 5,000 & 71.71 & 76.22 & 80.68 \\
13 & 10,000 & 72.02 & 78.29 & 82.17 \\
13 & 10,449 & 69.84 & 76.50 & 81.31 \\
2025 & 5,000 & 76.01 & -- & 79.64 \\
2025 & 10,000 & 70.71 & -- & 81.09 \\
2025 & 10,449 & 68.26 & -- & 80.07 \\
\bottomrule
\end{tabular}

 \caption{200M source-retrieval accuracy (percent) across training. Every entry uses 1,680 complete source-rotation groups. Seed-2025 SWA+Full was not evaluated.}
 \label{tab:probe_checkpoints}
\end{table}

\subsection{Distance-resolved source use}
\label{app:distance_effects}
\begin{figure}[htbp]
 \centering\includegraphics[width=\linewidth]{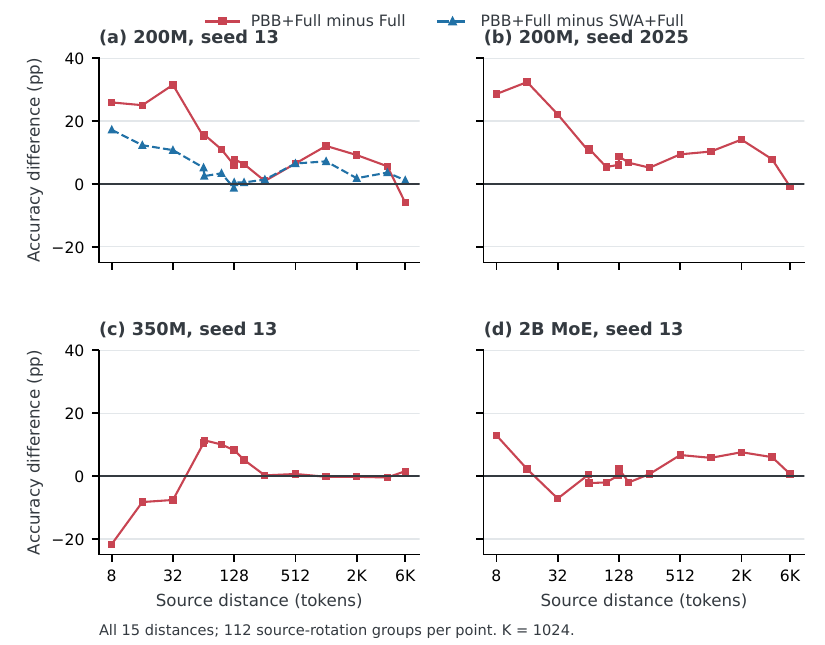}
 \caption{\textbf{Where the hybrid's source-retrieval gains occur.} Accuracy differences in percentage points on all fifteen measured distances; every point averages 112 matched source-rotation groups. Red compares PBB+Full with Full; blue additionally compares it with SWA+Full at 200M seed 13. Lines connect point estimates, without fitted interpolation or uncertainty bands. Positive and negative differences use the same vertical scale across panels.}
 \label{fig:distance_effects}
\end{figure}

Figure~\ref{fig:distance_effects} retains all fifteen distances of the frozen probe, averaging the seven phases and sixteen groups per phase: 112 groups at each distance. Positive differences favor PBB+Full. The figure separates the two 200M seeds, 350M, and MoE, and includes the seed-13 SWA hybrid comparison.

At 200M, aggregating distances up to 128 tokens gives PBB+Full advantages over Full of 17.22 and 15.60 percentage points for seeds 13 and 2025. The corresponding seed-13 advantage over SWA+Full is 6.28 points. Gains are not confined to nearby sources: over distances 2,048, 4,096, and 6,144, the MoE hybrid achieves 45.54\% accuracy versus Full's 40.77\%, with a source-gain increase of 0.227 nats. These are exploratory decompositions of existing measurements, not separately tuned probes. The aggregate does not imply improvement at every distance: the 200M Full comparison reverses at 6,144, 350M loses at the shortest distances, and MoE has mixed short-distance results. The complete curves make these differences explicit.

\subsection{Literal needle and Semantic exact retrieval}
\label{app:retrieval_families}
We report two additional controlled synthetic tasks, evaluated on the same final checkpoints as the main comparisons. \emph{Literal needle retrieval} inserts a statement such as \texttt{The secret word is red.} into natural held-out text and repeats its prefix at the readout. Four templates use ``secret word,'' ``recorded answer,'' ``key value,'' and ``code word.'' \emph{Semantic exact} measures exact entity--attribute retrieval: it inserts an attribute statement such as \texttt{Alice's ribbon was red.} and queries \texttt{Alice's ribbon was}. The four entity--object pairs are Alice--ribbon, David--scarf, Maria--ribbon, and James--scarf. Source and query wording match within each task; the latter measures exact entity--attribute retrieval, not paraphrase understanding.

Both tasks use distances
\[
 \{128,256,512,768,1024,1536,2048,2560,3072,3584,4096,4608,5120,6144,7168\}
\]
and readout phases $\{0,16,32,63,64,96,127\}$. At each distance--phase cell, 16 anchors cycle through the four templates or entity--object pairs. The readout is at $8064+r$, the source label at $8064+r-d$, and the stored input contains 8,192 tokens. Causal scoring uses only the prefix through the readout. A complete four-label rotation changes only the source label token; background, query, and positions remain fixed. Each task has 1,680 groups and 6,720 scored rows. The formal manifest was generated before these checkpoint evaluations, with generation seed 20260915, and shared unchanged across architectures. Accuracy and source gain $G$ follow the balanced-group definitions above. These tasks have their own distance grid and are reported separately from the original phase--distance probe.

\begin{table}[htbp]
 \centering\small\setlength{\tabcolsep}{4pt}
 \begin{tabular}{lllrrrr}
\toprule
 & & & \multicolumn{2}{c}{Literal needle} & \multicolumn{2}{c}{Semantic exact}\\
\cmidrule(lr){4-5}\cmidrule(lr){6-7}
Scale & Seed & Attention & Acc. & $G$ & Acc. & $G$\\
\midrule
200M & 13 & Full & 52.38 & 1.1450 & 53.32 & 1.5474\\
 & & Block & 25.00 & 0.0000 & 25.00 & 0.0000\\
 & & SWA & 25.19 & 0.0200 & 25.10 & 0.0044\\
 & & PBB & 25.24 & 0.0117 & 25.03 & 0.0028\\
 & & Random-Edge & 25.58 & 0.0276 & 25.46 & 0.0188\\
 & & SWA+Full & 53.65 & 1.3513 & 61.18 & 1.9106\\
 & & PBB+Full & 55.80 & 1.4202 & 58.07 & 1.9749\\
 & & WideBlock-192+Full & 55.95 & 1.4384 & 54.05 & 1.3320\\
\midrule
200M & 2025 & Full & 48.82 & 1.0522 & 52.77 & 1.5327\\
 & & PBB+Full & 55.03 & 1.3685 & 56.47 & 1.7375\\
\midrule
350M & 13 & Full & 50.04 & 1.1877 & 53.35 & 1.7135\\
 & & Block & 25.00 & 0.0000 & 25.00 & 0.0000\\
 & & PBB+Full & 50.85 & 1.2871 & 56.03 & 2.1348\\
\midrule
2B MoE & 13 & Full & 51.76 & 1.3469 & 52.44 & 2.0605\\
 & & PBB+Full & 56.88 & 1.6827 & 66.46 & 2.9172\\
\bottomrule
\end{tabular}

 \caption{\textbf{Literal needle and Semantic exact retrieval across all 15 final checkpoints.} Each task uses 6,720 balanced rows, with 25\% chance accuracy. Accuracies are percentages and $G$ is source gain in nats. Names ending in +Full use the 3:1 hybrid schedule. Values are point estimates; the original phase--distance probe is a separate evaluation.}
 \label{tab:retrieval_families}
\end{table}

PBB+Full improves both accuracy and source gain over Full on both tasks in all four matched comparisons: the two 200M seeds, 350M, and 2B MoE (Table~\ref{tab:retrieval_families}). At 2B MoE, needle accuracy increases by 5.12 percentage points with a 0.336-nat source-gain increase; Semantic exact accuracy increases by 14.02 points with a 0.857-nat gain increase. At 200M seed 13, WideBlock-192+Full and PBB+Full have close needle accuracy (55.95\% and 55.80\%), while PBB+Full has higher Semantic exact accuracy (58.07\% versus 54.05\%) and source gain (1.975 versus 1.332). SWA+Full has the highest Semantic exact accuracy among the 200M models, whereas PBB+Full has slightly higher source gain. Accuracy and source sensitivity thus give complementary views of the local graph.

\subsection{Paired context-window ablation}
\label{app:context}
The completed ablation uses 128 uniformly spaced, nonoverlapping 8K spans from the held-out test stream and the 200M seed-13 final checkpoints. All three architectures score the same 64 final target tokens of each span under contexts 128, 512, 2,048, and 8,192: 8,192 target tokens per architecture and context setting. These suffix readouts occupy phases 64--127, not the directly repaired first half-block. Truncated position origins reset, but the chosen context lengths preserve block phase, and RoPE relative offsets remain matched. The protocol records every corpus offset; it does not select spans based on architecture performance.

Table~\ref{tab:context} reports mean suffix NLL and excess relative to the same model's 8K context. Percentile intervals use 10,000 paired sequence-bootstrap replicates with seed 20260905. At 128 tokens, the PBB+Full minus SWA+Full penalty difference is 0.000289 nats, with interval $[-0.009844,0.010333]$. The corresponding difference against Full is $-0.013370$ nats, with interval $[-0.025428,-0.001357]$. PBB+Full achieves lower absolute suffix NLL at all four settings, with a truncation penalty similar to SWA+Full and smaller than Full at 128 tokens.

\begin{table}[htbp]
 \centering\small\begin{tabular}{lrrrr}
\toprule
Architecture & Context & Suffix NLL & Excess NLL & 95\% interval \\
\midrule
Full & 128 & 3.2012 & 0.2136 & [0.1759, 0.2550] \\
Full & 512 & 3.0263 & 0.0387 & [0.0225, 0.0570] \\
Full & 2,048 & 2.9908 & 0.0032 & [-0.0014, 0.0089] \\
Full & 8,192 & 2.9876 & 0.0000 & [0.0000, 0.0000] \\
SWA+Full & 128 & 3.1906 & 0.1999 & [0.1634, 0.2392] \\
SWA+Full & 512 & 3.0301 & 0.0394 & [0.0228, 0.0578] \\
SWA+Full & 2,048 & 2.9917 & 0.0010 & [-0.0028, 0.0055] \\
SWA+Full & 8,192 & 2.9907 & 0.0000 & [0.0000, 0.0000] \\
PBB+Full & 128 & 3.1784 & 0.2002 & [0.1645, 0.2389] \\
PBB+Full & 512 & 3.0168 & 0.0385 & [0.0218, 0.0575] \\
PBB+Full & 2,048 & 2.9827 & 0.0045 & [0.0002, 0.0098] \\
PBB+Full & 8,192 & 2.9782 & 0.0000 & [0.0000, 0.0000] \\
\bottomrule
\end{tabular}

 \caption{Matched context ablation, 128 test spans. Excess NLL is relative to each model's 8K suffix NLL; brackets give sequence-bootstrap 95\% intervals.}
 \label{tab:context}
\end{table}

\section{Checkpoint-Level Validation}
\label{app:curves}
Tables~\ref{tab:curve200}--\ref{tab:curvemoe} report synchronized validation trajectories under the shared warmup--stable--decay schedule. The curves show how each pair develops across training, while final held-out comparisons use equal-budget endpoints. Architecture pairs share the learning rate at each matched update; comparisons across updates also reflect the schedule.

\begin{figure}[htbp]
 \centering\includegraphics[width=\linewidth]{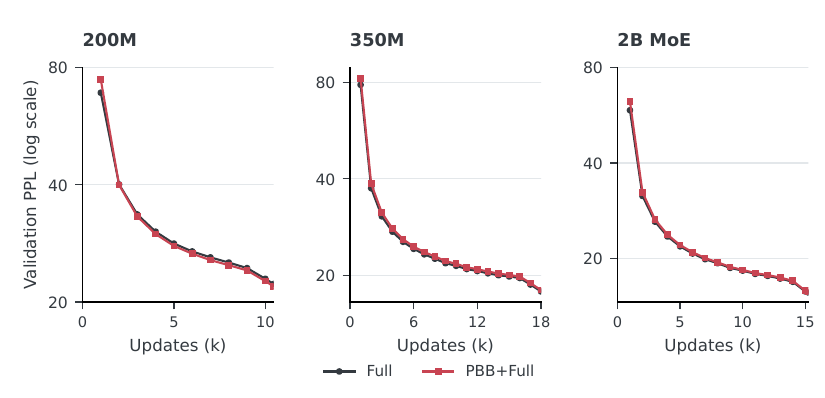}
 \caption{Synchronized seed-13 validation trajectories for Full and PBB+Full. Axes show model-specific update budgets ($k=1000$) and logarithmic PPL. The exact values and relative gaps are given in the following tables.}
\end{figure}

\begin{table}[htbp]
 \centering\small\begin{tabular}{rrrr}
\toprule
Step & Full PPL & PBB+Full PPL & Gap (\%) \\
\midrule
1,000 & 68.94 & 74.56 & 8.14 \\
2,000 & 40.11 & 40.00 & -0.28 \\
3,000 & 33.57 & 33.16 & -1.21 \\
4,000 & 30.33 & 29.89 & -1.43 \\
5,000 & 28.27 & 27.88 & -1.39 \\
6,000 & 26.99 & 26.62 & -1.34 \\
7,000 & 26.05 & 25.63 & -1.62 \\
8,000 & 25.27 & 24.88 & -1.54 \\
9,000 & 24.46 & 24.12 & -1.38 \\
10,000 & 22.95 & 22.63 & -1.38 \\
10,449 & 22.26 & 21.95 & -1.39 \\
\bottomrule
\end{tabular}

 \caption{200M, seed 13: synchronized validation PPL. Gap is $100(\mathrm{PPL}_{\PBB+\Full}/\mathrm{PPL}_{\Full}-1)$.}
 \label{tab:curve200}
\end{table}
\begin{table}[htbp]
 \centering\small\begin{tabular}{rrrr}
\toprule
Step & Full PPL & PBB+Full PPL & Gap (\%) \\
\midrule
1,000 & 78.55 & 82.34 & 4.83 \\
2,000 & 37.36 & 38.58 & 3.27 \\
3,000 & 30.53 & 31.35 & 2.69 \\
4,000 & 27.34 & 27.98 & 2.33 \\
5,000 & 25.43 & 25.91 & 1.92 \\
6,000 & 24.18 & 24.56 & 1.59 \\
7,000 & 23.24 & 23.62 & 1.62 \\
8,000 & 22.52 & 22.90 & 1.68 \\
9,000 & 21.81 & 22.14 & 1.52 \\
10,000 & 21.40 & 21.71 & 1.48 \\
11,000 & 20.91 & 21.19 & 1.32 \\
12,000 & 20.61 & 20.88 & 1.30 \\
13,000 & 20.28 & 20.54 & 1.27 \\
14,000 & 20.00 & 20.24 & 1.23 \\
15,000 & 19.82 & 20.05 & 1.14 \\
16,000 & 19.61 & 19.82 & 1.08 \\
17,000 & 18.70 & 18.90 & 1.06 \\
18,000 & 17.80 & 17.97 & 0.93 \\
\bottomrule
\end{tabular}

 \caption{350M, seed 13: synchronized validation PPL under the same definition.}
 \label{tab:curve350}
\end{table}
\begin{table}[t]
 \centering\small\begin{tabular}{rrrr}
\toprule
Step & Full PPL & PBB+Full PPL & Gap (\%) \\
\midrule
1,000 & 58.68 & 62.50 & 6.50 \\
2,000 & 31.48 & 32.35 & 2.74 \\
3,000 & 26.12 & 26.53 & 1.58 \\
4,000 & 23.45 & 23.75 & 1.27 \\
5,000 & 21.84 & 22.04 & 0.93 \\
6,000 & 20.73 & 20.92 & 0.94 \\
7,000 & 19.88 & 20.06 & 0.89 \\
8,000 & 19.32 & 19.44 & 0.65 \\
9,000 & 18.66 & 18.80 & 0.74 \\
10,000 & 18.31 & 18.41 & 0.54 \\
11,000 & 17.87 & 17.98 & 0.59 \\
12,000 & 17.62 & 17.73 & 0.61 \\
13,000 & 17.28 & 17.41 & 0.74 \\
14,000 & 16.88 & 17.00 & 0.69 \\
15,000 & 15.73 & 15.82 & 0.52 \\
15,259 & 15.61 & 15.69 & 0.50 \\
\bottomrule
\end{tabular}

 \caption{2B MoE, seed 13: synchronized validation PPL. Both models receive 6.000B tokens.}
 \label{tab:curvemoe}
\end{table}

\section{Kernel, Training, and Cached-Inference Protocols}
\label{app:efficiency}
\subsection{Phase-aligned support and exact ring storage}
\label{app:ring_cache}
For $b=2h$, Equation~\eqref{eq:pbb_relay} gives the oldest accessible position
$a(t)=\max(0,h(\lfloor t/h\rfloor-1))$. It is nondecreasing and changes only at half-block boundaries. After the initial block, the number of sources, including the current token, is
\begin{equation}
 |\A_{\PBB}(t)|=h+1+(t\bmod h)\leq 2h=b.
 \label{eq:pbb_cache_support}
\end{equation}
Its steady-state phase average is $(3h+1)/2$, or 96.5 at $h=64$, versus SWA's 128 sources. This count describes logical reads; physical allocation and executed tile work are separate quantities.

\begin{proposition}[Relocation-free PBB cache]
Under sequential, single-token causal decoding with $b=2h$, storing position $s$ in slot $s\bmod b$ preserves every KV entry required by the current and all future PBB readouts. The ring requires $b$ entries per KV head and no relocation of retained entries.
\end{proposition}
\paragraph{Proof.}
At position $t=p+r\geq b$, writing slot $t\bmod b$ overwrites position $t-b$. If $r<h$, then $t-b=p+r-2h<p-h=a(t)$. If $r\geq h$, then $t-b<p=a(t)$. Thus the overwritten position lies strictly before the current support. Since $a(t)$ never decreases, no future readout requires it. Current sources form an interval of at most $b$ positions, so their slots are distinct. Initial-block writes do not overwrite valid entries. These facts establish the claim by induction. Absolute RoPE positions remain unchanged by the physical slot mapping.

At the default width, bridge phases read the preceding tail in slots 64--127 followed by the current prefix in slots 0--63; the other phases read the current prefix contiguously. Flash-PBB implements these layouts directly in a 128-column, four-warp FlashAttention-derived kernel, fusing KV append and attention without staging gathers or buffer compaction. Native GQA groups query heads over shared KV heads. The implementation retains the compatible low-precision softmax/reduction order; it does not add another normalization or approximate the support.

For batch $B$, $H_{\rm KV}$ KV heads, head dimension $d_h$, and element size $u$, physical local KV storage is $2BbH_{\rm KV}d_hu$. With $B=1$, $b=128$, $H_{\rm KV}=2$, $d_h=64$, and BF16, this is 65,536 bytes. The previous Flash-PBB and tested Flash-SWA paths allocate 256 positions, or 131,072 bytes, and relocate 65,536 bytes of retained KV payload every 128 steps. The ring removes this relocation, equivalent to 512 payload bytes per token per layer averaged over phases. Payload is an array-copy count, not a measured DRAM transaction count. SWA also admits a 128-entry ring; the measured allocation reduction is relative to the tested backend, not an intrinsic SWA storage lower bound.

\subsection{Same-device Flash-PBB decode comparison}
\label{app:flash_decode}
The new Flash-PBB, previous Flash-PBB, and Flash-SWA are compared on the same otherwise-idle rental device, using batch 1, BF16, two KV heads, head dimension 64, and 10/14/12 query heads for the 200M/350M/2B shapes. Eight absolute prefix positions span 8K--1M. These positions do not create a growing local cache or require whole-model 1M prefill. Each shape/position is tested in three independent processes, totaling 72 cases; all pass the recorded no-external-SM checks.

\begin{figure}[htbp]
 \centering\includegraphics[width=\linewidth]{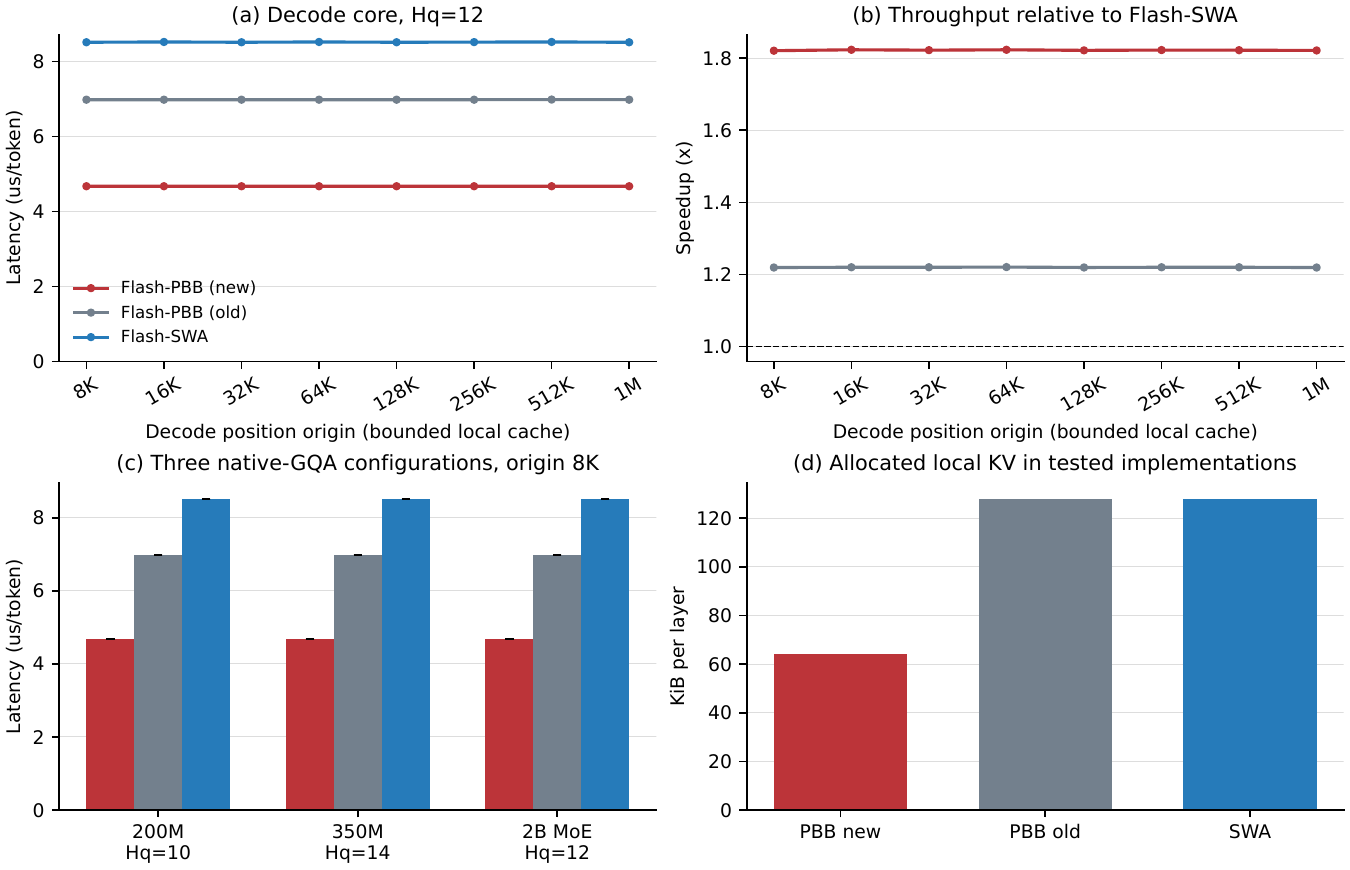}
 \caption{\textbf{Decode implementation comparison.} New Flash-PBB, previous Flash-PBB, and Flash-SWA under the same device and protocol. Local KV is physical allocation; timing includes KV maintenance. Prefix positions label bounded-cache tests, not full-model context capacity.}
 \label{fig:flash_decode_history}
\end{figure}

Each process measures nine 256-step CUDA-Graph sequences per backend, with balanced cyclic timing order. A sequence visits every local phase twice. Timing includes attention, KV append, and amortized rollover maintenance, but excludes projections, RoPE, experts, and vocabulary output. New/old PBB outputs first match bitwise over 384 sequential steps. Each backend separately passes eager/graph checks for the last output and full cache contents. Figure~\ref{fig:flash_decode_history} retains the old implementation as an engineering baseline; the main figure compares the new kernel with Flash-SWA.

Across all 24 shape/position cells, new Flash-PBB/Flash-SWA throughput ratios range from $1.8188\times$ to $1.8235\times$. At the 2B shape and 8K origin, median step times are 4.672, 6.980, and 8.512\,$\mu$s for new PBB, old PBB, and SWA. Local KV allocation is 64, 128, and 128\,KiB, respectively. The kernel update affects decode and its cache layout; it is not the source of the separate training/prefill gains.

\subsection{Measured execution resources and memory requests}
\label{app:decode_resources}
We profile the same compiled SM89 kernels on a local RTX 4060 Laptop GPU using Nsight Compute 2024.1.1, Torch 2.6.0+cu124, and FlashAttention 2.8.3.post1. The synthetic BF16 inputs use batch 1, 12 query heads, two KV heads, and head dimension 64, matching the MoE attention shape. At position origin 8K, chronological decode samples phases $\{0,20,63,64,80,127\}$, covering bridge entry, bridge exit, and maximum support. New/old PBB outputs match bitwise over 384 steps.

Each backend is run in three independent processes per cache policy, with rotating backend order: 18 runs and 108 profiled attention calls. Cold-cache collection uses kernel replay with cache flushing; retained-cache collection uses application replay without flushing after identical warmup. The profiler selects only the attention kernel, including fused KV append but excluding separate rollover copies. Figure~\ref{fig:flash_decode}b and Table~\ref{tab:decode_resources} report the measured resources and memory requests.

\begin{table}[htbp]
 \centering\small\setlength{\tabcolsep}{4pt}
 \begin{tabular}{lrrr}
\toprule
Metric & Flash-PBB & Previous PBB & Flash-SWA\\
\midrule
Registers/thread & 212 & 255 & 255\\
Dynamic shared memory/block (KiB) & 40 & 72 & 72\\
Physical local KV (KiB) & 64 & 128 & 128\\
Global-load sectors/call & 1,112--2,120 & 1,112--2,120 & 15,168\\
Global-store sectors/call & 67 & 67 & 156\\
Thread-local load sectors/call & 0 & 32 & 2,880\\
Thread-local store sectors/call & 0 & 32 & 5,952\\
\bottomrule
\end{tabular}

 \caption{\textbf{Lower resource use and fewer memory requests.} RTX 4060, 2B attention shape. Counts repeat exactly across all three processes and both cache policies at each phase; ranges span the six sampled phases, not uncertainty intervals. Registers and shared memory are launch resources; sector counts are hardware-observed requests.}
 \label{tab:decode_resources}
\end{table}

Flash-PBB uses 212 registers/thread and 40\,KiB dynamic shared memory/block, versus 255 and 72\,KiB for Flash-SWA: reductions of 16.9\% and 44.4\%. All kernels use 128 threads/block; PBB launches two query-group blocks and SWA twelve query-head blocks. These resource attributes also match the independent rental-device launch traces. The new binary has zero stack and static local-memory allocation.

Global-load requests fall from SWA's 15,168 sectors/call to PBB's 1,112--2,120, a reduction of 86.0--92.7\% at the sampled phases. The counter is \texttt{l1tex\_\_t\_sectors\_pipe\_lsu\_mem\_global\_op\_ld.sum}; it counts sector requests, including repeated requests, rather than unique source bytes. New PBB also records zero thread-local load/store requests, compared with 2,880/5,952 for SWA. Thread-local memory here is CUDA thread-private storage, not the attention KV cache; the SWA disassembly contains corresponding \texttt{LDL/STL} stack accesses, which are absent from the new PBB kernel.

New and previous PBB have identical global-load requests within attention. The new implementation instead reduces register/shared-memory allocation, removes the previous kernel's 32/32 thread-local load/store requests, and eliminates rollover relocation. The smaller tile, grouped dispatch, and ring addressing jointly produce the tested implementation; the request reduction versus SWA is not explained by edge count alone. The packaged observations preserve every phase and repeat, and regenerate the figure without a GPU.

\paragraph{L2 read and write diagnostics.}
L2 counters also favor Flash-PBB: median read-sector counts are lower at all six phases under both cache policies. Median write-sector counts are 80--81 for Flash-PBB versus 6,121 for Flash-SWA. Table~\ref{tab:decode_l2} retains each phase's median and full three-run range. These are \texttt{lts\_\_t\_sectors\_op\_read.sum} and \texttt{lts\_\_t\_sectors\_op\_write.sum}, distinct from the L1/TEX requests in Figure~\ref{fig:flash_decode}. Because chip-global L2 counters also include concurrent desktop activity, this table is diagnostic rather than an isolated estimate of kernel traffic or DRAM-bandwidth savings.

\begin{table}[htbp]
 \centering\scriptsize\setlength{\tabcolsep}{3pt}
 \begin{tabular}{llrrrr}
\toprule
Cache & Phase & PBB reads & SWA reads & PBB writes & SWA writes\\
\midrule
Cold & 0 & 8,063 [7,813, 16,485] & 24,262 [22,759, 41,837] & 80 [80, 80] & 6,121 [6,121, 6,172]\\
Cold & 20 & 8,288 [8,006, 8,338] & 27,228 [23,372, 41,698] & 80 [80, 80] & 6,121 [6,121, 6,121]\\
Cold & 63 & 8,881 [8,703, 9,410] & 23,228 [23,078, 24,195] & 80 [80, 80] & 6,121 [6,121, 6,121]\\
Cold & 64 & 8,063 [7,968, 8,080] & 28,953 [26,571, 32,765] & 80 [80, 80] & 6,121 [6,121, 6,172]\\
Cold & 80 & 8,316 [8,204, 16,669] & 26,572 [22,737, 27,933] & 80 [80, 80] & 6,121 [6,121, 6,121]\\
Cold & 127 & 9,012 [8,704, 11,249] & 26,957 [22,858, 30,435] & 80 [80, 80] & 6,121 [6,121, 6,121]\\
Retained & 0 & 7,977 [7,867, 9,344] & 23,284 [22,773, 27,508] & 80 [80, 80] & 6,121 [6,121, 6,121]\\
Retained & 20 & 7,992 [7,990, 8,247] & 22,696 [22,692, 27,731] & 81 [80, 144] & 6,121 [6,121, 6,122]\\
Retained & 63 & 8,822 [2,214, 10,307] & 25,832 [22,732, 27,235] & 80 [80, 80] & 6,121 [6,121, 6,168]\\
Retained & 64 & 7,976 [7,669, 15,513] & 22,718 [22,691, 22,720] & 80 [80, 80] & 6,121 [6,121, 6,121]\\
Retained & 80 & 7,918 [1,469, 9,431] & 29,668 [22,712, 36,323] & 80 [80, 80] & 6,121 [6,121, 6,163]\\
Retained & 127 & 8,762 [8,681, 9,986] & 22,696 [22,690, 30,192] & 80 [80, 80] & 6,121 [6,121, 6,122]\\
\bottomrule
\end{tabular}

 \caption{\textbf{L2 sector counts during profiled attention: Flash-PBB versus Flash-SWA.} RTX 4060, 2B attention shape. Each entry is median [minimum, maximum] over three runs; all phases and both cache policies are shown. Chip-global diagnostic scope, including background activity.}
 \label{tab:decode_l2}
\end{table}

\subsection{Exact union and numerical checks}
The added PBB edges form $64\times64$ target-by-source rectangles: 64 post-boundary queries each read 64 sources from the previous block. The fused kernel visits causal block tiles and these rectangles without constructing a dense sequence-square mask. Projections and KV heads are shared with the original GQA layer. Each source enters the single-softmax normalization once.

The RTX 4090 BF16 forward check compares PBB against exact union references at lengths 128, 256, and 1,024. Maximum output error is at most 0.015625, with cosine similarity above 0.999997. Its length-128 backward check covers the initial block: gradient cosine similarities exceed 0.999994, with maximum errors up to 0.03125. The multi-block check below additionally tests bridge gradients. At 8K, a separate LSE-specialization forward check reports identical outputs and finite log-partition values.

A supplementary multi-block check exercises the grouped4 backward implementation on an RTX 4060 Laptop GPU with Torch 2.11.0+cu128 and Triton 3.6.0. It uses batch 1, 12 query heads, two KV heads, head dimension 64, and BF16 inputs, with an independently constructed FP32 dense-union reference. Two random tensor seeds cover lengths 128, 256, and 1,024. At both multi-block lengths, additional cases restrict the upstream gradient to repaired queries 128--191 and inspect gradients of previous-block keys and values at positions 64--127. All ten cases pass the prescribed relative-$L_2$ threshold of 0.02 and cosine threshold of 0.9998. Across outputs and $dQ,dK,dV$, the largest relative error is 0.002562 and the smallest cosine is 0.9999966; maximum absolute $dV$ error is 0.03741. The previous-block gradients have nonzero reference norm and relative error below 0.002416. The experiment archive retains the kernel snapshot, executable check, and per-case results for this correctness experiment.

These BF16-versus-reference checks cover initial blocks, empty bridges, causal masks, and native GQA at the stated numerical tolerances. The CPU algebra checks separately verify 1,980 graph/depth configurations, the disjoint LSE identity, and exact PBB/WideBlock score equality.

The new training sweep uses the grouped4 PBB backward path and Triton grouped experts. All three architectures use the same compatibility adjustment for Triton 3.2: a loop-invariant tile-count constant is declared outside the expert loop. Sparse/empty-expert forward and backward checks precede timing; no architecture-specific change is made to expert arithmetic.

\subsection{Shared hardware and scope}
The primary execution matrix uses RTX 4090 (48\,GB) GPUs with matched software and phase-specific benchmark settings. Per-run hardware records are retained in the accompanying numerical artifact. The model has 2,069,563,136 stored and 300,091,136 active parameters, 20 layers, hidden width 768, 12 query heads, two KV heads, and head dimension 64. Full has 20 Full layers; hybrids have 15 local and five Full layers at zero-based indices 3, 7, 11, 15, and 19. SWA has window 128; PBB has block 128 and bridge-exclusive width 64. All paths use native GQA without repeating KV heads.

The execution benchmarks use seeded, randomly initialized weights and synthetic device-resident inputs to isolate architecture execution from data I/O. Trained MoE routing and serving requests can produce different expert loads. Quality is evaluated separately on trained checkpoints at 8K; the longer-context sweeps measure execution speed and capacity under the specified runtime.

The rental sweeps use Torch 2.6.0+cu124, Triton 3.2.0, Transformers 5.10.2, and FlashAttention 2.8.3.post1, with eight CPU threads for whole-model runs and four for the new decode-core comparison. Processes start separately for each case. All 33 successful whole-model cells, including seven updated PBB inference cells, pass the recorded no-external-SM checks. Full/SWA and training data are unchanged; only PBB inference is replaced. Table~\ref{tab:eff_protocols} defines the protocols. Earlier laboratory-card measurements appear separately in Appendix~\ref{app:legacy}.

\begin{table}[htbp]
 \centering\scriptsize\setlength{\tabcolsep}{3pt}
 \begin{tabular}{@{}p{0.15\linewidth}p{0.26\linewidth}p{0.25\linewidth}p{0.26\linewidth}@{}}
 \toprule
 Protocol & State / expert backend & Work per measurement & Repetition\\
 \midrule
 Context training & FP32 masters, BF16 AMP; Triton grouped experts & Micro-batch 1; complete update including clearing & Two warmups; five updates\\
 Cached prefill & BF16; grouped-matrix-multiply expert dispatch & Full input, KV population, final-position logits & Three warmed calls at every successful length\\
 Sequential decode & BF16; Triton grouped experts; CUDA Graphs for all models & 256 incremental tokens, advancing positions and KV state & Three alternating eager/graph pairs; graph times reported\\
 Attention layer & BF16; native GQA; kernel and complete-layer scopes & Prefill or 256 sequential layer steps & Five prefill calls and five decode continuations per scope\\
 \bottomrule
 \end{tabular}
 \caption{Primary efficiency protocols on RTX 4090 (48\,GB). AMP denotes automatic mixed precision. Each phase uses matched settings across architectures. Reported medians and min--max ranges summarize within-process repetitions.}
 \label{tab:eff_protocols}
\end{table}

\subsection{Complete-update context sweep}
The sweep fixes 131,072 input tokens per optimizer update and micro-batch 1 through 128K. At contexts 8K, 16K, 32K, 64K, and 128K, gradient accumulation is 16, 8, 4, 2, and 1. The 256K, 512K, and 1M attempts use one sequence per update as a separate capacity group. This fixed-batching protocol matches architectures at each length while exposing how context length changes attention cost and accumulation passes.

Parameters, gradients, and fused-AdamW moments are FP32 and GPU-resident. Computation uses BF16 autocast, non-reentrant activation checkpointing, the same Triton grouped MoE implementation, and fused vocabulary loss. AdamW uses learning rate $3\times10^{-4}$, betas $(0.9,0.95)$, $\epsilon=10^{-8}$, weight decay 0.1, and gradient clipping at norm 1.0. Seeded initialization uses seed 13. Each timed update includes gradient clearing, forward computation, language-model and router auxiliary losses, accumulated backward, clipping, and the optimizer update. Device synchronization brackets timing; setup, first-use compilation, data allocation, and validation are excluded. Synthetic inputs are reused across accumulation passes.

Two complete warmup updates precede five measured updates. Table~\ref{tab:training_context} reports their median and observed range, totaling 60 timed updates across twelve successful cases. Speedup is the ratio of architecture and Full median throughputs at the same length.

\begin{table}[htbp]
 \centering\scriptsize\setlength{\tabcolsep}{3pt}
 \begin{tabular}{@{}llrrrr@{}}
\toprule
Context & Architecture & GA & k tok/s [min, max] & s/update & Peak GiB \\
\midrule
8K & Full & 16 & 9.15 [9.06, 9.15] & 14.33 & 32.93 \\
8K & SWA+Full 3:1 & 16 & 9.79 [9.79, 9.80] & 13.38 & 32.93 \\
8K & PBB+Full 3:1 & 16 & 9.74 [9.68, 9.75] & 13.45 & 32.93 \\
16K & Full & 8 & 12.01 [11.98, 12.02] & 10.91 & 33.93 \\
16K & SWA+Full 3:1 & 8 & 13.90 [13.76, 13.93] & 9.43 & 33.93 \\
16K & PBB+Full 3:1 & 8 & 13.96 [13.82, 13.98] & 9.39 & 33.93 \\
32K & Full & 4 & 12.21 [12.20, 12.21] & 10.74 & 36.38 \\
32K & SWA+Full 3:1 & 4 & 16.58 [16.55, 16.61] & 7.90 & 36.38 \\
32K & PBB+Full 3:1 & 4 & 16.66 [16.64, 16.67] & 7.87 & 36.38 \\
64K & Full & 2 & 9.74 [9.74, 9.74] & 13.46 & 41.28 \\
64K & SWA+Full 3:1 & 2 & 16.61 [16.60, 16.62] & 7.89 & 41.28 \\
64K & PBB+Full 3:1 & 2 & 16.70 [16.69, 16.72] & 7.85 & 41.28 \\
128K & All three architectures & 1 & OOM & -- & -- \\
256K & All three architectures & 1 & OOM & -- & -- \\
512K & All three architectures & 1 & OOM & -- & -- \\
1M & All three architectures & 1 & OOM & -- & -- \\
\bottomrule
\end{tabular}

 \caption{Complete training-update sweep, all three architectures. GA is gradient accumulation; k tok/s uses $k=1000$. Successful cells each contain five measured updates. All 128K--1M attempts fail: 128K during the timed stage and the longer cases during warmup; no failed cell contributes a throughput estimate.}
 \label{tab:training_context}
\end{table}

At 64K, PBB+Full, SWA+Full, and Full reach 16.703k, 16.614k, and 9.738k tokens/s, respectively. PBB/Full is $1.715\times$, while PBB/SWA is $1.005\times$. Peak allocated memory is 41.283\,GiB for all three, dominated by parameter, optimizer, and activation storage. All 128K cases complete two warmup updates, then fail on a 768\,MiB allocation during the timed stage; each longer attempt also exhausts memory. Under this configuration, 64K is the largest context completing all five timed updates. Training memory covers the full process; persistent KV storage is measured separately in inference.

\subsection{Exact sequential graph decoding}
The cached runner uses matched initialization with seed 1309, BF16 weights/computation, and batch 1. Prefill creates KV state and returns last-position logits using the same grouped-matrix-multiply expert dispatch for all architectures. Before decode, every architecture switches to the same Triton grouped expert implementation. Each incremental step updates the cache, executes all layers, projects the output vocabulary, and takes a GPU argmax. No full-prefix recomputation is performed. PBB local layers use 128-slot rings; the tested SWA backend retains 256-position buffers. Full layers retain the input plus 385 positions reserved for checks and continuations. Allocation does not change either local pattern's logical support.

Prefill uses FlashAttention for Full/SWA and fused Triton for PBB. Decode uses the FlashAttention KV-cache API for Full/SWA and Flash-PBB's exact ring layout for PBB. A bank of 128 phase-specific CUDA Graphs handles local geometry and updates. Absolute RoPE positions and Full-cache lengths advance on device, and each graph consumes the preceding step's selected token. Graph replay includes projections, all expert layers, cache updates, output logits, and argmax, implementing the complete sequential generation step.

Every successful whole-model case checks 384 sequential positions against eager execution with the same low-precision expert backend. All full-vocabulary logits match bitwise, and the final absolute position and Full-cache length are explicitly checked. The span covers every local phase and three block rollovers, validating graph execution and state updates within each architecture.

\subsection{Cached timing and capacity results}
Table~\ref{tab:cached} reports all 24 requested cached cases. After a separately recorded cold prefill, each successful case measures three warmed cache-populating calls. The repetition count is identical from 8K through 512K. Graph capture, correctness checks, state restoration, and cold-start work are excluded from steady-state throughput.

Following the graph correctness check, each case measures three alternating eager/graph pairs; the table reports the three graph continuations. Each restores the prefix state and generates the same greedy trajectory of 256 successive tokens. CPU wall time is bracketed by device synchronization; CUDA-event timings are retained separately. The context labels exclude the continuation. All three architectures complete 8K, 16K, 32K, 64K, 128K, 256K, and 512K. The archived numerical records retain cold times, all repeats, parity outcomes, and memory scopes.

\begin{table}[htbp]
 \centering\scriptsize\setlength{\tabcolsep}{2.3pt}
 \begin{tabular}{@{}llrrrr@{}}
\toprule
Context & Architecture & Prefill k/s [min, max] & Decode/s [min, max] & Peak GiB & KV GiB \\
\midrule
8K & Full & 106.63 [105.25, 107.75] & 329.81 [326.02, 332.30] & 4.24 & 0.082 \\
8K & SWA+Full 3:1 & 110.55 [108.21, 110.90] & 334.32 [334.00, 340.66] & 4.18 & 0.022 \\
8K & PBB+Full 3:1 & 111.18 [110.80, 111.24] & 348.86 [346.98, 352.56] & 4.18 & 0.021 \\
16K & Full & 93.96 [93.10, 94.13] & 303.67 [302.67, 306.55] & 4.61 & 0.160 \\
16K & SWA+Full 3:1 & 118.00 [114.32, 118.04] & 330.99 [330.45, 333.66] & 4.49 & 0.042 \\
16K & PBB+Full 3:1 & 118.07 [116.31, 119.77] & 343.64 [337.62, 345.86] & 4.49 & 0.041 \\
32K & Full & 71.14 [71.07, 71.18] & 285.78 [284.54, 289.13] & 5.35 & 0.316 \\
32K & SWA+Full 3:1 & 105.84 [105.76, 105.85] & 325.08 [322.13, 326.70] & 5.12 & 0.081 \\
32K & PBB+Full 3:1 & 106.39 [105.49, 106.89] & 335.16 [333.28, 337.64] & 5.12 & 0.080 \\
64K & Full & 50.09 [49.91, 50.11] & 258.02 [257.79, 260.48] & 6.84 & 0.629 \\
64K & SWA+Full 3:1 & 92.49 [92.43, 92.54] & 313.25 [312.59, 314.77] & 6.37 & 0.159 \\
64K & PBB+Full 3:1 & 93.39 [93.38, 93.44] & 320.58 [318.48, 323.06] & 6.37 & 0.158 \\
128K & Full & 31.26 [31.26, 31.39] & 209.05 [208.46, 209.46] & 9.81 & 1.254 \\
128K & SWA+Full 3:1 & 73.27 [73.20, 73.32] & 290.99 [290.40, 294.31] & 8.87 & 0.315 \\
128K & PBB+Full 3:1 & 73.87 [73.85, 73.91] & 302.32 [301.12, 303.83] & 8.87 & 0.314 \\
256K & Full & 17.64 [17.59, 17.71] & 158.71 [158.41, 158.92] & 15.75 & 2.504 \\
256K & SWA+Full 3:1 & 50.67 [50.61, 50.73] & 264.37 [263.57, 265.73] & 13.87 & 0.628 \\
256K & PBB+Full 3:1 & 51.18 [51.08, 51.23] & 269.81 [269.66, 273.71] & 13.87 & 0.627 \\
512K & Full & 9.13 [9.13, 9.13] & 105.11 [105.09, 105.12] & 27.63 & 5.004 \\
512K & SWA+Full 3:1 & 30.39 [30.34, 30.44] & 218.34 [217.43, 218.83] & 23.88 & 1.253 \\
512K & PBB+Full 3:1 & 30.31 [30.21, 30.47] & 221.87 [221.58, 223.74] & 23.88 & 1.252 \\
1M & Full & OOM & -- & -- & -- \\
1M & SWA+Full 3:1 & OOM & -- & -- & -- \\
1M & PBB+Full 3:1 & OOM & -- & -- & -- \\
\bottomrule
\end{tabular}

 \caption{Cache-populating prefill and sequential graph decode, complete context matrix. Successful rates are medians over three measurements, with observed ranges; decode continuations contain 256 tokens. KV is physical cache storage, excluding graph workspace. OOM denotes an attempted prefill that exhausted memory.}
 \label{tab:cached}
\end{table}

At 512K, PBB+Full records 30.307k prefill tokens/s versus Full's 9.130k ($3.319\times$). Median sequential decode is 221.873 versus 105.106 tokens/s ($2.111\times$). SWA+Full records 30.386k prefill and 218.340 decode tokens/s. PBB/SWA hybrids allocate 1.25183/1.25275\,GiB of KV versus Full's 5.00367\,GiB, because only five layers retain the long global cache (Table~\ref{tab:cached}). Halving PBB's 15 local buffers saves 983,040 additional bytes, small beside the global cache at this length.

At 1M ($1{,}048{,}576$) input tokens, Full fails during cold prefill on a 5.00\,GiB allocation with 2.56\,GiB free. Both hybrids fail on a 6.00\,GiB allocation, with 4.11\,GiB free for SWA and 4.13\,GiB for PBB. Whole-model execution completes through 512K; the isolated-layer tests do not imply whole-model 1M capacity. Decode-process peaks include the graph pool and reference buffers in addition to model state.

\subsection{Execution time per token}
For measured throughput $v$ tokens/s, the steady-state GPU time per million tokens is $10^6/(3600v)$ hours. The primary throughput measurements can therefore be converted directly into execution-time costs. At 512K, PBB+Full reduces prefill time per token by 69.9\% and decode time per token by 52.6\% relative to Full. The 64K training-update ratio corresponds to 41.7\% less time at fixed tokens/update. These are steady-state costs under the specified protocols; a request's combined cost depends on its input and output lengths.

\section{Earlier Bridge-Family Experiments}
\label{app:legacy}
The development experiments below examine residual-branch Bridge models and frozen-checkpoint mask interventions. Their model families, fusion rules, and prompt protocols differ from the main experiments; their exploratory probe results precede the balanced source-rotation design in Appendix~\ref{app:probes} and are reported separately. The earlier training-recipe benchmark is retained under its original batching and timing protocol.

\subsection{124M, 1024-token controlled training}
The earlier GPT-2-like setting used learned absolute positions, a GPT-2 tokenizer, OpenWebText, 1,024-token context, and three training seeds \citep{radford2019language,gokaslan2019openwebtext}. Bridge-family layers used separately normalized residual branches with shared projections. With block and auxiliary outputs $o_1,o_2$, the layer computes $W_O(o_1+o_2)$, so a source in both supports participates in both normalizations. PBB limits auxiliary write-back to the post-boundary half; SE-Bridge extends the previous-block source range. Table~\ref{tab:legacy124} gives the reported aggregate measurements.

\begin{table}[htbp]
 \centering\small
 \begin{tabular}{lrrrr}
 \toprule
 Attention & Validation PPL & Needle & Bridge-window & Clean semantic\\
 \midrule
 Full &28.56&0.967&0.989&0.946\\
 Block &36.79&0.429&0.415&0.392\\
 Bridge &31.74&0.437&0.812&0.420\\
 PBB &31.38&0.438&0.842&0.451\\
 SE-Bridge &30.84&0.436&0.983&0.516\\
 SWA &30.56&0.513&0.991&0.493\\
 \bottomrule
 \end{tabular}
 \caption{Earlier 124M/1K results under the residual-branch training and probe protocol. Bridge-family entries use shared-projection, separately normalized branches.}
 \label{tab:legacy124}
\end{table}

SWA led the local models on language modeling and standard needle retrieval \citep{kamradt2023needle}, while wider-source SE-Bridge recovered the boundary-window deficit. Clean semantic retrieval was 0.516 for SE-Bridge and 0.493 for SWA at the evaluated checkpoints. This pattern motivated separating source coverage, write-back geometry, and fusion in subsequent experiments.

\subsection{Frozen Qwen2.5-7B, 8K hybrid intervention}
The earlier 7B experiment imposed masks on a pretrained Qwen2.5 base checkpoint, retaining its weights and GQA projections without fine-tuning. Every fourth layer remained Full, and sparse PBB/SE-Bridge layers used a single-softmax union implemented with a dense mask. Evaluation ranked candidates on 832 prompts; the overall score in Table~\ref{tab:legacy7b} weights task families by their displayed example counts.

\begin{table}[htbp]
 \centering\small
 \begin{tabular}{lrrrrr}
 \toprule
 Probe & Examples & Block+Full & SWA+Full & PBB+Full & SE+Full\\
 \midrule
 Needle &64&0.313&0.547&0.578&0.578\\
 Bridge-window &312&0.458&0.583&0.599&0.590\\
 Semantic boundary &312&0.401&0.471&0.455&0.446\\
 Semantic exact &48&0.375&0.521&0.479&0.500\\
 Paraphrase &48&0.313&0.500&0.458&0.458\\
 Distractor &48&0.000&0.167&0.146&0.104\\
 \midrule
 Weighted overall &832&0.386&0.506&0.502&0.494\\
 \bottomrule
 \end{tabular}
 \caption{Earlier 7B frozen-checkpoint hybrid intervention, 3:1 local/full. Displayed values are rounded from the original result table.}
 \label{tab:legacy7b}
\end{table}
The frozen-checkpoint comparison separates literal/boundary retrieval from semantic behavior: PBB+Full leads SWA+Full on needle and bridge-window retrieval, while SWA+Full leads on the semantic variants. Their task-weighted aggregate accuracies are close, with gains concentrated in different task families.

\subsection{Earlier 8K recipe benchmark}
The earlier optimized benchmark uses the training-run batch configuration: micro-batch 2, accumulation 24, and 393,216 tokens/update. Both architectures use Triton grouped experts, FP32 parameters/gradients/Adam moments, BF16 autocast, activation checkpointing, and no offload or state sharding. The runtime is Torch 2.11.0+cu128 with FlashAttention 2.8.3.post1. Gradient clearing occurs outside this timed region; forward, loss, accumulated backward, clipping, and optimizer update are included.

After one warmup, Full update times are 24.8275, 26.0002, and 31.6251 seconds; PBB+Full times are 22.3821, 22.4117, and 22.4044 seconds. Their medians yield 15,123.6 and 17,550.8 tokens/s (+16.05\%). Both record 33.932\,GiB peak allocated and 35.072\,GiB reserved memory. This batching configuration is reported separately from the micro-batch-1 context sweep in Figure~\ref{fig:efficiency}d.

\end{document}